\documentclass[journal]{IEEEtran}
\usepackage{amsmath,amsfonts}
\usepackage{amsthm,amssymb}
\usepackage{mathrsfs}

\usepackage{algorithmic}
\usepackage{algorithm}
\usepackage{array}
\usepackage[caption=false,font=normalsize,labelfont=sf,textfont=sf]{subfig}
\usepackage{textcomp}
\usepackage{stfloats}
\usepackage{url}
\usepackage{verbatim}
\usepackage{graphicx}
\usepackage{cite}

\usepackage{soul}
\usepackage{bm}
\usepackage{booktabs}
\usepackage{txfonts}
\usepackage{booktabs}
\usepackage{multirow}
\usepackage{caption3}
\usepackage{xcolor}

\usepackage{listings}
\usepackage{pifont}
\usepackage{threeparttable}

\begin{document}

\title{Change-Agent: Towards Interactive Comprehensive Remote Sensing Change Interpretation and Analysis 
}
% Towards User-Interactive Comprehensive Interpretation from Remote Sensing Change Detection and Change Captioning
% Advancing from Change Captioning and Detection towards User-Interactive Change Interpretation and Processing
% User-interactive Remote Sensing Change Interpretation Agent: Bridge Change Captioning and Change Detection
% 从变化检测和变化描述到带有任务联合的变化解译智能体
% 利用两个任务之间的关联打造强大的变化解译智能体（智能体的特点就是可交互，不用明说了）
% From Remote Sensing Image Change Captioning and Change Detection to  User-interactive Change Interpretation Agent
% Joint Learning for Remote Sensing Change Captioning and Change Detection: A Dataset and a Transformer-based Method
% 从变化检测和变化captioning到变化解译智能体
% Comprehensive Remote Sensing Image Change Interpretation: Joint Learning for Change Captioning and Change Detection
% "Unified Learning Framework for Change Detection and Description in Remote Sensing Images"
% "Joint Change Detection and Description in Dual-Temporal Remote Sensing Images"
% 介绍数据集时：
% 1、不同物体的尺度：介绍目标建筑物和道路的大小
% 2、形状：

% Chenyang Liu, ~\IEEEmembership{Graduate Student Member,~IEEE},
% \author{Chenyang Liu, Keyan Chen, Jianqi Chen, Haotian Zhang, Zipeng Qi, Hao Chen,~\IEEEmembership{Member,~IEEE}, \\ Zhengxia Zou, and Zhenwei Shi$^*$,~\IEEEmembership{Senior Member,~IEEE}
\author{Chenyang Liu, Keyan Chen, Haotian Zhang, Zipeng Qi, Zhengxia Zou,~\IEEEmembership{Member,~IEEE}, \\ and Zhenwei Shi$^*$,~\IEEEmembership{Senior Member,~IEEE}
\\
\vspace{8pt}
Beihang University
% \thanks{The work was supported by the National Key Research and Development Program of China (Grant No. 2022ZD0160401), the National Natural Science Foundation of China under Grant 62125102, the Beijing Natural Science Foundation under Grant JL23005, and the Fundamental Research Funds for the Central Universities. \emph{(Corresponding author: Zhenwei Shi (e-mail: shizhenwei@buaa.edu.cn))}
% }
\thanks{Chenyang Liu, Keyan Chen, Haotian Zhang, Zipeng Qi, and Zhenwei Shi are with the Image Processing Center, School of Astronautics, with the Beijing Key Laboratory of Digital Media, and with the State Key Laboratory of Virtual Reality Technology and Systems, Beihang University, Beijing 100191, China, also with the Shanghai Artificial Intelligence Laboratory, Shanghai 200232, China.

Chenyang Liu is also with Shen Yuan Honors College of Beihang University, Beijing 100191, China.

Zhengxia Zou is with the Department of Guidance, Navigation and Control, School of Astronautics, Beihang University, Beijing 100191, China, and also with Shanghai Artificial Intelligence Laboratory, Shanghai 200232, China.
}
}



\maketitle

\begin{abstract}
Monitoring changes in the Earth's surface is crucial for understanding natural processes and human impacts, necessitating precise and comprehensive interpretation methodologies. Remote sensing satellite imagery offers a unique perspective for monitoring these changes, leading to the emergence of remote sensing image change interpretation (RSICI) as a significant research focus. Current RSICI technology encompasses change detection and change captioning, each with its limitations in providing comprehensive interpretation. To address this, we propose an interactive Change-Agent, which can follow user instructions to achieve comprehensive change interpretation and insightful analysis, such as change detection and change captioning, change object counting, change cause analysis, etc. The Change-Agent integrates a multi-level change interpretation (MCI) model as the eyes and a large language model (LLM) as the brain. The MCI model contains two branches of pixel-level change detection and semantic-level change captioning, in which the BI-temporal Iterative Interaction (BI3) layer is proposed to enhance the model's discriminative feature representation capabilities. To support the training of the MCI model, we build the LEVIR-MCI dataset with a large number of change masks and captions of changes. Experiments demonstrate the SOTA performance of the MCI model in achieving both change detection and change description simultaneously, and highlight the promising application value of our Change-Agent in facilitating comprehensive interpretation of surface changes, which opens up a new avenue for intelligent remote sensing applications. To facilitate future research, we will make our dataset and codebase publicly available at \emph{\url{https://github.com/Chen-Yang-Liu/Change-Agent}}

\end{abstract}


\section{Introduction}
% influenced by both natural evolution and human activities,
% With the rapid advancements in remote sensing technology, r
\IEEEPARstart{T}{he} changes in the Earth's surface, as a dynamic indicator of the Earth's system, profoundly affect the planet's evolution and the survival of humankind. Observing and analyzing these changes is pivotal for advancing sustainable human development \cite{lv2022land_CD_survey}. Remote sensing (RS) satellite imagery, offering a unique ``God's perspective," emerges as an effective tool for monitoring Earth's dynamic changes. Remote sensing image change interpretation (RSICI) has emerged as a significant research focus \cite{Zheng_202410439252}. RSICI aims to identify and analyze changes from images captured at different times in the same geographical area. It can provide decision-making support for environmental protection \cite{Noman_10489990}, urban planning \cite{Lin_10415489}, resource management \cite{wang2024CD_review}, etc.

Current RSICI technology primarily encompasses change detection and change captioning. Change detection accurately localizes the spatial location of surface changes at the pixel level \cite{ISPRS2022changemask,CD_feng2023change}. In contrast, change captioning aims at articulating the attributes and meanings of changes using natural language, emphasizing the semantic-level understanding of changes \cite{RSICC_1,RSICC_2,RSICCformer,RSICC_TIP2023,liu2023decoupling}. Despite significant advancements in both areas, their respective limitations prevent us from obtaining comprehensive change interpretation information by utilizing a single technology. Specifically, although change detection can accurately localize the changed area, {it lacks a deeper semantic-level understanding of changes' underlying meanings, such as the characteristics of ground objects (e.g., category, color, shape), the spatial relationships between them (e.g., ``beside'', ``around'') and dynamics of changes (e.g., ``appear'', ``removed'')}. Conversely, change captioning can provide rich semantic-level interpretation information but may fall short in providing precise pixel-level change localization. Therefore, there is a pressing need to explore a multi-level change interpretation (MCI) approach providing both pixel-level and semantic-level change information, facilitating precise change localization while delving into the essence and implications of the changes.

Moreover, practical applications often require comprehensive analysis and further processing of both pixel-level and semantic-level change interpretation results to meet specific needs. For instance, users may require statistical analysis of changed object numbers, consuming significant time and effort from researchers and demanding users' technical proficiency. 

\begin{figure*}
	\centering
	\includegraphics[width=1\linewidth]{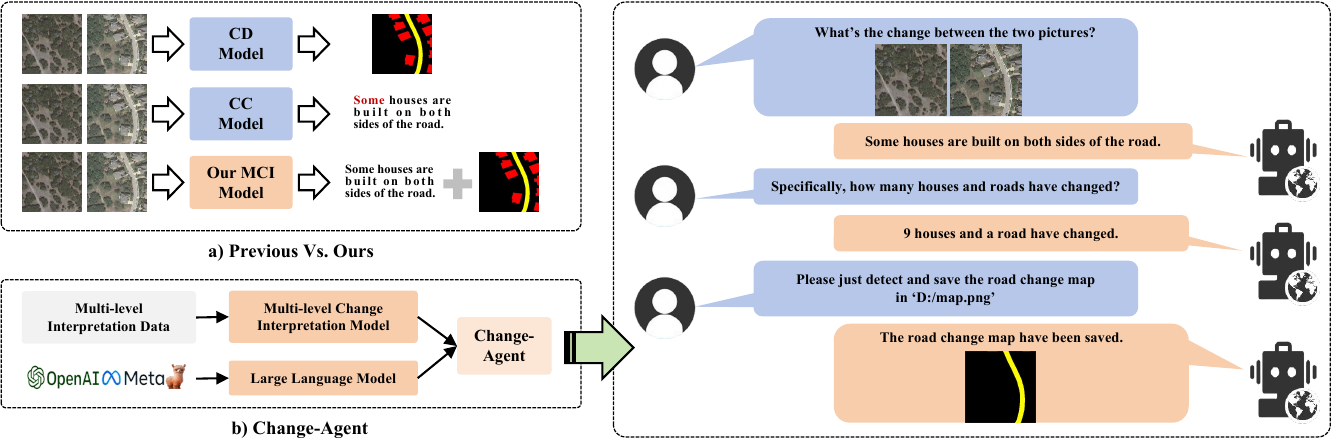}
        \vspace{-15pt}
	\caption{{The comparison between previous single technology and our Change-Agent. Our Change-Agent can simultaneously achieve precise pixel-level change detection and semantic-level change captioning. Besides, it boasts interactive capabilities, enabling users to communicate their queries on surface changes.}
 }
	\label{fig:idea}
\vspace{-10pt}
\end{figure*}

To address these challenges, we propose an interactive Change-Agent based on a novel MCI model and a large language model (LLM). Fig.\ref{fig:idea} illustrates the advancements achieved by our Change-Agent compared to previous technologies. Different from the previous single technology, with the help of the MCI model, our Change-Agent can simultaneously achieve precise pixel-level change detection and semantic-level change captioning, thereby providing users with comprehensive change interpretation information. Notably, our Change-Agent boasts interactive capabilities, enabling users to communicate their queries or requirements regarding surface changes. Leveraging the MCI model and the LLM, the Change-Agent can adeptly interpret changes according to user needs, intelligently analyze and process the change interpretation information, and ultimately deliver tailored outcomes aligning with user expectations. The interactive Change-Agent bridges the gap between users and remote sensing expertise.

As illustrated in Fig.\ref{fig:idea} (b), the MCI model is a pivotal cornerstone toward the realization of the Change-Agent. It enables the Change-Agent to perceive visual changes comprehensively, serving as the change perception foundation of the Change-Agent. However, there has been no prior work investigating the multi-task learning of change detection and change captioning. To address this gap, we built the LEVIR-MCI dataset. Each pair of bi-temporal remote sensing images in this dataset contains both pixel-level change detection masks and semantic-level change descriptions. Building upon this dataset, we propose a novel MCI model with two branches of change detection and change captioning. Within the two branches, we propose BI-temporal Iterative Interaction (BI3) layers utilizing Local Perception Enhancement (LPE) and the Global Difference Fusion Attention (GDFA) modules to enhance the model's discriminative feature representation capabilities. The trained MCI model equips our Change-Agent with eyes, enabling the agent to realize comprehensive change interpretations of surface changes.

Recent advancements in LLMs such as ChatGPT \cite{openai2022chatgpt} and Llama2 \cite{llama2} have been remarkable. Trained on vast corpora, LLMs have acquired extensive knowledge and demonstrated powerful capabilities in instruction comprehension, planning, reasoning, question answering, and text generation \cite{Pretrained_model,prompt_survey}. These notable progress have provided robust support for our LLM-based Change-Agent. In this work, the LLM serves as the brain of the Change-Agent, assuming a pivotal role in internal scheduling, planning, and user-intent understanding. Integration with LLMs imbues our agent with enhanced flexibility and intelligence, enabling a nuanced understanding of user intentions and provision of customized change interpretation and intelligent analysis services such as changed object counting, estimation of causes, prediction of changes, etc.

{Our Change-Agent facilitates accurate and comprehensive change interpretation of RS images, offering novel perspectives for analyzing changes on the Earth's surface. It reduces the workload and time costs for researchers while enhancing the efficiency and convenience of interaction between users and remote sensing data. Extensive experiments demonstrate the effectiveness of the proposed MCI model and highlight the promising application value of our Change-Agent.}
% Besides, the integration of the MCI model and LLM presents novel perspectives for analyzing surface changes.
% and intelligent remote sensing applications.
 % presents novel ideas and approaches for intelligent remote sensing applications, offering 
% presents novel ideas and approaches for intelligent remote sensing applications, offering fresh perspectives for analyzing changes on the Earth's surface. 

% 我们提出的数据集以及关于多任务学习模型为未来的变化解译提供了新的方向。我们的
% 变化检测和变化描述的多任务学习和数据集以及智能体的探索为未来的研究提供了
Our contributions can be summarized as follows:
\begin{itemize}
\item 
We develop an interactive Change-Agent, which uses external tools to achieve comprehensive interpretation and analysis of surface changes according to the user's instructions. It has intelligent dialogue and customized service capabilities, opening a new avenue for intelligent remote-sensing applications.

% Leveraging the MCI model and an LLM, we construct a Change-Agent that achieves an interactive and comprehensive interpretation and analysis of surface changes. It has intelligent dialogue and customized service capabilities, opening up new opportunities for intelligent remote sensing applications.
% Equipped with intelligent dialogue and customized service capabilities, this agent paves the way for new avenues in remote sensing intelligence applications.

\item 
% 为了给智能体配备强大的全面的变化解译能力，我们提出了一个基于多任务学习的包含变化检测和变化描述双分支的变化解译模型。该模型能同时提供像素级和语义级的变化解译信息。经过多任务学习训练得到的两个分支将作为智能体的视觉感知工具。
% To equip the Change-Agent with comprehensive change interpretation capabilities, 
We propose a multi-task learning-based MCI model with dual branches for change detection and change captioning. This model can simultaneously provide pixel-level and semantic-level change interpretation information, serving as the perception tool of the Change-Agent.

% We propose a dual-branch MCI model, which can provide pixel-level and semantic-level change interpretation. Besides, we propose BI3 layers with LPE and GDFA to enhance the change interpretation capability of the model. Experiments validate the effectiveness of our method.
% , facilitating comprehensive change interpretation of the Change-Agent

\item 
We built an MCI dataset named LEVIR-MCI, which contains bi-temporal images, corresponding change detection masks and descriptive sentences. The dataset provides a crucial data foundation for exploring multi-task learning for change detection and change captioning.
% advancing comprehensive change interpretation methodologies and 
\end{itemize}

\section{Related work}
A comprehensive change interpretation model should provide both pixel-level and semantic-level change information. However, previous work has primarily focused on either change detection or change captioning alone. Change detection aims to generate pixel-level change maps revealing the locations of changes in bi-temporal images, while change captioning aims to describe the changes via language. 
% In this section, we will briefly review these approaches.

\subsection{Remote Sensing Change Detection}
Early change detection approaches were traditionally categorized into algebra-based, transformation-based, and classification-based methods. These methods relied on techniques such as change vector analysis (CVA) \cite{CVA_1,CVA_2}, principal component analysis (PCA) \cite{deng2008pca,CD_PCA2}, multivariate alteration detection (MAD) \cite{MAD_Nielsen,MAD_Nielsen2007}, and post-classification comparisons \cite{post_classification_1,post_classification_2}. 
Traditional methods have laid the foundation for modern change detection techniques \cite{surey_1}. 

With the emergence of deep learning\cite{chen2023TTP,chen2022contrastive,chen2024digital,liu2022hyperspectral,liu2022physics,liu2023diverse}, the field of remote sensing change detection has witnessed significant advancements \cite{li2022densely_CD,wu2023fully,chen2024rsmamba,BIFA,chen2023rsprompter,chen2022degraded,chen2023dense, chen2021building}. Compared to traditional methods, deep learning significantly improves the performance of change detection with its superior multi-temporal feature learning capability. These methods utilize various architectures including convolutional neural networks (CNNs) \cite{FC-Siam, Zheng_202410439252,lv2023multi_CD}, recurrent neural networks (RNNs) \cite{LSTM, Conv-LSTM,GRU,CD_LSTM_202410502016,SiamCRNN}, auto-encoders \cite{AE_2006_Hinton,kingma2013auto_ae,vincent2008extracting}, and Transformers \cite{BIT,CD_Ding_202410443352,TransUNetCD,Lin_10415489,ding2024joint,qi2023implicit}. 
For instance, Chen \textit{et al.} \cite{BIT} introduced a bi-temporal image Transformer (BIT) \cite{BIT}, where images are represented as semantic tokens, and the Transformer encoder and decoder refine visual tokens to identify changed areas efficiently and effectively. Li \textit{et al.} \cite{TransUNetCD} proposed an end-to-end TransUNetCD model based on Transformer and UNet structure. The model utilizes a Transformer encoder to extract global contextual features. A CNN-based decoder upsamples the multi-scale features to generate the change map. Bandara \textit{et al.} proposed a pure transformer model named ChangeFormer \cite{changeformer}, whose backbone is a siamese SegFormer \cite{xie2021segformer}. Zhang \textit{et al.} \cite{CD_zhang2023remote} propose a deep siamese network with multi-scale multi-attention. They enhance global representation using a contextual attention module and fuse multi-scale features from the siamese feature extractor for detecting objects of varying sizes and irregularities. Compared to binary change detection, semantic change detection considers the semantic information of change categories, which is presented in the form of semantic masks. Some works, such as \cite{yang2021asymmetric,zhu2022land}, have significantly advanced the development of semantic change detection. However, higher-level semantic information, such as the relationships between changed objects, their colors, and spatial distribution, remains challenging to directly obtain. Some surveys, such as \cite{CDSurvey_DL, wang2024CD_review} provide a more comprehensive review of change detection in deep learning.

% To improve change detection accuracy, many recent works have been developed for improving the feature representation ability and the change discrimination ability of the model, such as proposing different feature fusion strategies \cite{ multi_level_feature_fusion_1,multi_level_feature_fusion_2,multi_level_feature_fusion_3,multi_level_feature_fusion_4,CD_attention_3,urban_1}, designing different attention modules \cite{urban_1,CD_attention_1,CD_attention_2,CD_attention_3}, and introducing the self-attention mechanism \cite{urban_1,BIT}. For instance, \cite{CD_attention_2} proposed a pyramid feature-based attention-guided Siamese network (PGA-SiamNet), in which a co-attention module captures the correlation between bi-temporal images and a context fusion strategy fuses low-level and high-level features.

\subsection{{Remote Sensing Change Captioning}}
Remote sensing change captioning (RSCC) is a recently emerged multimodal task involving remote sensing image processing and natural language generation. Given the intricacy of this task, prevailing methodologies predominantly leverage deep learning techniques. Early methods in this domain largely drew inspiration from the encoder-decoder framework commonly utilized in the image captioning task \cite{Yang_202410415446,Meng_202410492984,Liu_2022}. Within this framework, the encoder is responsible for extracting features from bi-temporal images and performing the difference-aware fusion of these features. The decoder typically employs recurrent neural networks or Transformers to convert visual features into natural language. Compared to tasks like image captioning, which extract semantic representations from single images, change description faces the significant challenge of extracting robust semantic change representations.

Chouaf and Hoxha \textit{et al.} \cite{RSICC_1, RSICC_2} pioneered remote sensing change captioning by introducing a model that employs a pre-trained CNNs as the encoder and tried to utilize an RNN and Support Vector Machines (SVMs) as the decoder to generate textual descriptions. To improve the change-awareness of the model, they proposed an image-based fusion strategy and a feature-based fusion strategy. To facilitate research on change captioning, Liu \textit{et al.} \cite{RSICCformer} introduced a large LEVIR-CC dataset and benchmarked some methods from other fields \cite{robust_CC,MCCformer}. Besides, they proposed a Transformer-based method named RSICCformer. The approach incorporates Siamese cross-encoding modules and multistage bitemporal fusion modules, emphasizing change regions through differencing features and capturing multiple changes of interest. 

In a further advancement, some attention-based improvements have emerged \cite{RSICC_TIP2023,RSICC_cai2023RS_interactive,liu2024rscama,PSNet,liu2023pixel,10485459}. For instance, PSNet \cite{PSNet} enhances the perception of changed objects of varying sizes using a progressive scale-aware network with difference perception layers and scale-aware reinforcement modules. Chang \textit{et al.} \cite{RSICC_TIP2023} proposed an attention network containing a hierarchical self-attention module followed by a residual unit for identifying change-related attributes and constructing the semantic change embedding. Unlike the encoder-decoder framework, Liu \textit{et al.} \cite{liu2023decoupling} proposed a decoupling paradigm by designing an image-level classifier and a feature-level encoder. Besides, a multi-prompt learning strategy is proposed to effectively exploit an LLM for captioning.

\begin{figure}
	\centering
	\includegraphics[width=1.0\linewidth]{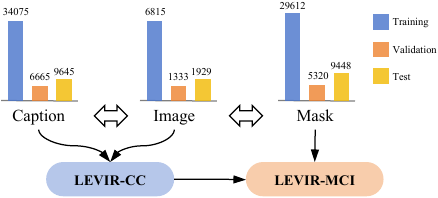}
	\caption{{The comparison of the proposed LEVIR-MCI dataset and the previous LEVIR-CC dataset. The LEVIR-MCI dataset is an extension of LEVIR-CC. 
 }
 }
 % \vspace{-10pt}
\label{fig:data_compare}
\end{figure}

\section{LEVIR-MCI Dataset}
To cultivate a robust MCI model, foundational to the Change-Agent, we propose an MCI dataset named LEVIR-MCI (LEVIR Multi-level Change Interpretation) \footnote{LEVIR is our laboratory name: the Learning, Vision, and Remote Sensing Laboratory.}, featuring both pixel-level change information in the form of change detection masks and semantic-level insights encapsulated in descriptive sentences. The LEVIR-MCI dataset contains 10077 bi-temporal images. Each image has a spatial size of 256 × 256 pixels with a high resolution of 0.5 m/pixel and has a corresponding annotated mask and five annotated captions. The dataset will be publicly available at: \emph{\url{https://github.com/Chen-Yang-Liu/Change-Agent}}

The proposed LEVIR-MCI dataset is an extension of our previously established change captioning dataset, LEVIR-CC \cite{RSICCformer}. {As shown in} Fig. \ref{fig:data_compare}, we further provided each pair of bi-temporal images with additional change detection masks highlighting changed roads and changed buildings. Unlike its predecessor, the LEVIR-MCI dataset provides diverse annotations from different interpretation perspectives for each pair of images, further enhancing its utility for comprehensive change interpretation. 

\begin{figure*}
	\centering
	\includegraphics[width=1.0\linewidth]{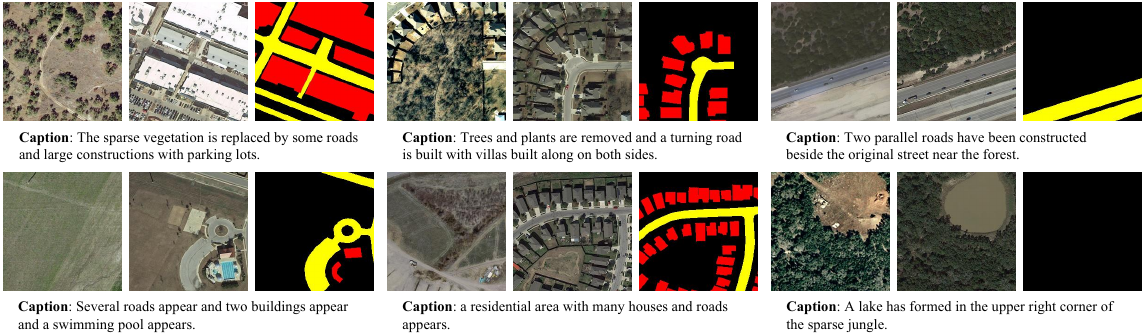}
         % \vspace{-5pt}
	\caption{{Examples of the LEVIR-MCI dataset. Each pair of bi-temporal images is provided with a change detection mask and one of the five sentences describing changes. In the change detection mask, changed buildings are highlighted in red, while changed roads are depicted in yellow.
 } 
 }
	\label{fig:dataset}
\end{figure*}

Fig. \ref{fig:dataset} presents some examples from our proposed LEVIR-MCI. Each pair of bi-temporal images within the dataset is annotated by a change detection mask delineating pixel-level alterations, along with one of five descriptive sentences elucidating the semantic-level changes. 
% In the change detection mask, changed buildings are highlighted in red, while changed roads are depicted in yellow. 
% Each sentence provides valuable insights into the evolving landscape.
% We can see that the two annotations provide change interpretation information from different angles. 
% The two annotation information are complementary, which is very important for building a powerful and comprehensive change interpretation model.
% , with pixel-level and semantic-level information mutually reinforcing each other. 
Such complementary annotations provide a multifaceted view of change interpretation, thereby bolstering the comprehensiveness of change interpretation models. To facilitate a nuanced understanding of the dataset, and considering our previous research has analysed the change captions, we will elucidate additional insights into the annotated change detection masks.

% from the number and scale of changed objects.
\begin{figure*}
	\centering
	\includegraphics[width=1.0\linewidth]{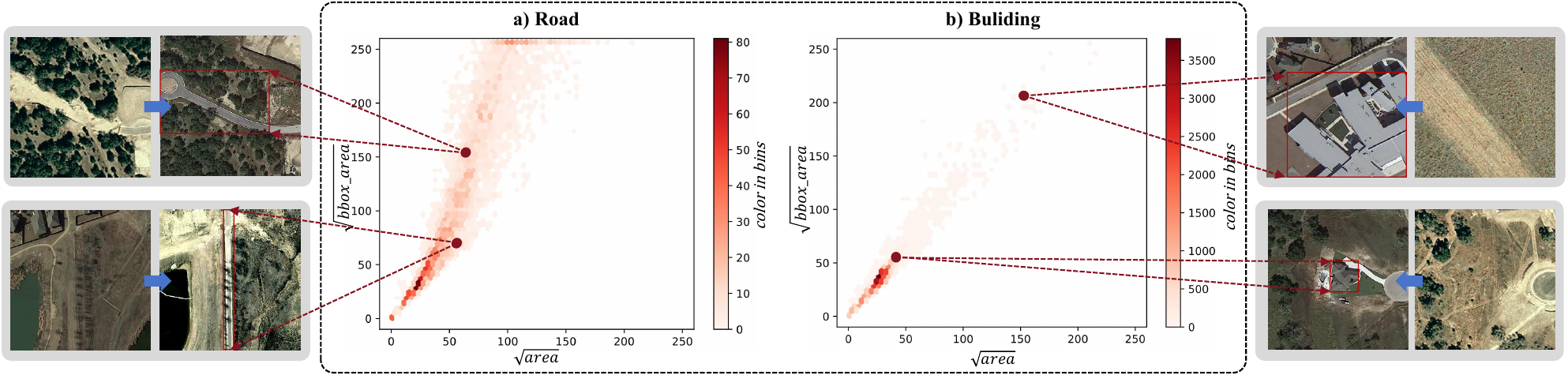}
 % \vspace{-5pt}
	\caption{Distribution of the scale and deformation of changed roads and buildings. The ``area'' on the horizontal axis represents the area of a single object, and the ``bbox\_area'' on the vertical axis represents the area of the corresponding rectangular bounding box. {The value of the color bins represents the number of object instances.} The dispersion of points offers insights into the diversity of object scale and deformation.
 }
	\label{fig:road_building}
\end{figure*}
\begin{table*}[!t] 
\renewcommand{\arraystretch}{1.1}
\caption{The Number of changed roads and buildings in the LEVIR-MCI dataset. The dataset comprises over 40,000 annotated instances of changed roads and buildings.}
\label{tab:number_object}
\centering
\begin{tabular}{c|c c c| c c c |c c c}
	\toprule%[1pt]
	% \midrule
	\multirow{2}{*}{Set} & \multicolumn{3}{c|}{Number of changed objects } & \multicolumn{3}{c|}{Number of images with changed objects} & \multicolumn{3}{c}{Average number of objects per image} \\ \cline{2-3} \cline{4-5} \cline{6-7} \cline{8-10}
    & & Road & Building & & Road & Building & & Road & Building\\
    \midrule
    {Training-set} & & 3457 & 26155 & & 2364 & 2972 & & 1.46 & 8.80\\
    
    {Validation-set} & & 611 & 4709 & & 424 & 567 & & 1.44 & 8.31\\
    {Test-set} & & 934 & 8514 & & 603 & 878 & & 1.55 & 9.70\\
    \midrule
    {Total dataset} & & 5002 & 39378 & & 3391 & 4417 & & 1.48 & 8.92 \\

    \midrule
\end{tabular}
\end{table*}

\subsection{Number of Changed Objects}
In Table \ref{tab:number_object}, we meticulously analyse changed object masks in the LEVIR-MCI dataset. The dataset comprises over 40,000 annotated instances of changed roads and buildings. While the number of changed roads is fewer than buildings, generally exhibit longer spans and cover larger areas, as evidenced in Fig. \ref{fig:road_building}. With such a vast collection of annotated masks, our LEVIR-MCI dataset can also serve as a fertile data ground for developing innovative multi-category change detection methodologies.

% This substantial volume of labeled data not only underscores the richness and diversity of our dataset but also accentuates its pivotal role in advancing the field of change detection. By offering a vast array of annotated instances, our dataset also can serve as a fertile ground for the exploration and development of innovative change detection methodologies.

\subsection{Scale and Deformation of Changed Objects}
\label{subsection:Scale} 
An insightful examination of object scale and deformation is conducted through an analysis of road and building areas and the corresponding rectangular bounding boxes, as illustrated in Fig. \ref{fig:road_building}. The ``area'' on the horizontal axis represents the area of a single object, and the ``bbox\_area'' on the vertical axis represents the area of the corresponding bounding box.
The dispersion of points offers insights into the diversity of object scale and deformation. Notably, points of roads exhibit a relatively broader spectrum with dispersed diversity, attributable to their narrow and curved nature. In contrast, points of buildings display a more concentrated distribution, reflecting their predominantly rectangular form. This analysis elucidates the intricate interplay between object scale and shape, shedding light on the nuanced characteristics of changed objects.
% , and paving the way for more refined and accurate change detection methodologies.

\begin{figure*}
	\centering
	\includegraphics[width=1\linewidth]{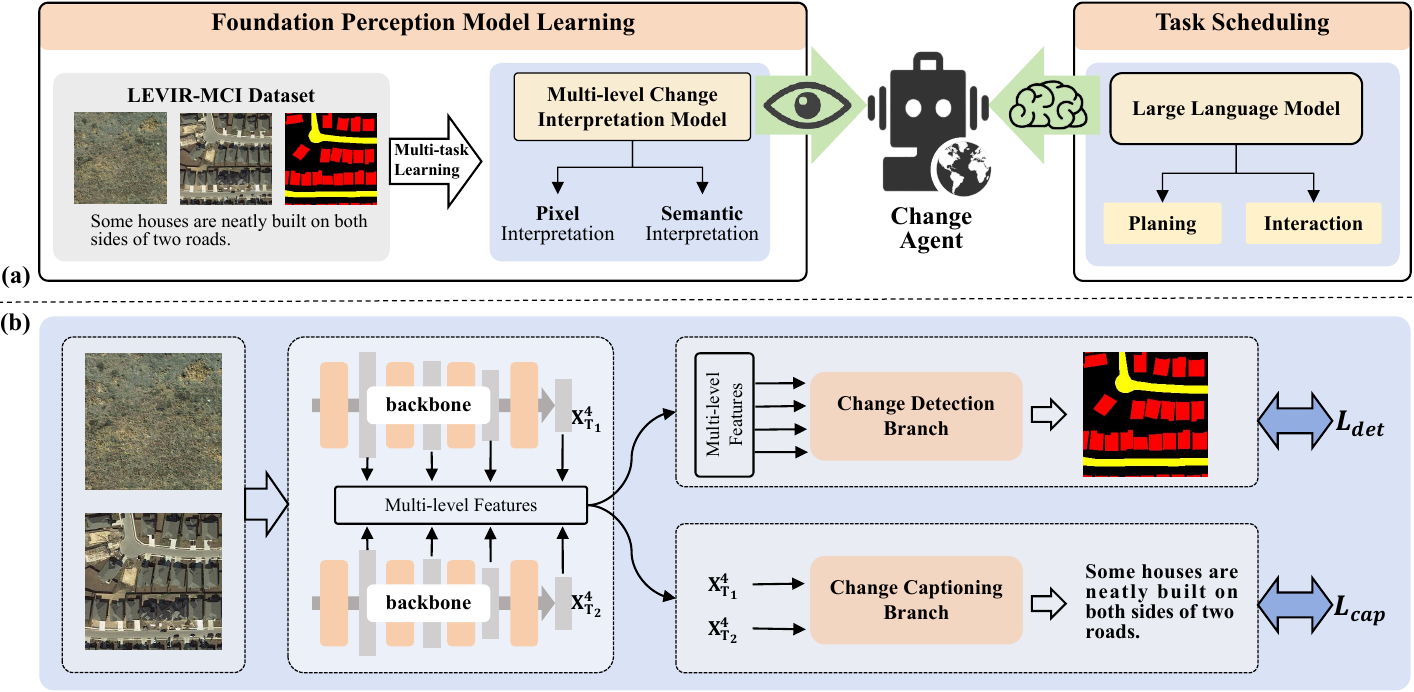}
	\caption{The overview of Change-Agent is shown in (a). The Change-Agent is equipped with an MCI model and an LLM, serving as its eyes and brain, respectively. The proposed LEVIR-MCI dataset provides a data foundation for training the MCI model. (b) shows the overall structure of the MCI model.
 }
	\label{fig:overall}
\end{figure*}

% In Fig. \ref{fig:overall}, we present an overview of the proposed Change-Agent. Comprising an MCI model as eyes and an LLM as the brain, the Change-Agent is endowed with visual perception and cognitive capabilities. This integration enables the agent to achieve an interactive and comprehensive interpretation and analysis of surface changes.
% a comprehensive understanding and interpretation of surface changes, empowering the agent to provide insightful analyses and decision-making support. The MCI model facilitates robust and comprehensive change interpretation capabilities, while the LLM orchestrates the planning and information processing aspects of the agent. In the subsequent sections, we delve into the construction and training of the MCI model and elucidate the LLM-based scheduling of the Change-Agent.

\section{METHODOLOGY}
In Fig. \ref{fig:overall}, we present the overview of the proposed Change-Agent. We utilize the MCI model as eyes and an LLM as the brain to build our Change-Agent. The MCI model enables the Change-Agent to perceive visual changes comprehensively, serving as the change perception foundation of the Change-Agent. The constructed LEVIR-MCI dataset supports the multi-task training of the MCI model, which contains a change detection branch and a change captioning branch. As another pivotal role of the agent, the LLM can use its inherent rich knowledge to implement agent scheduling and provide insightful analyses and decision-making support. In the subsequent section, we will introduce the composition of the MCI model and the LLM-based scheduling of the agent.

% is responsible for the planning and information processing of the Change-Agent and can use it . 
% offering insightful analyses and decision-making support

\subsection{Multi-level Change Interpretation Model}
The MCI model serves as a central component of the Change-Agent, tasked with extracting and interpreting change information from bi-temporal remote sensing images. The proposed LEVIR-MCI dataset lays the groundwork for training the MCI model. The MCI model adopts a dual-branch architecture with a shared bottom, focusing on two pivotal tasks: change detection and change captioning. Specifically, a Siamese backbone network extracts multi-scale visual features from bi-temporal images, enabling learning in two branches. The lower-level features offer detailed information, while higher-level features are rich in semantics. Our change detection branch utilizes multi-scale features to refine change mask predictions, while the change captioning branch leverages the highest-level visual features to generate descriptive sentences. Through multi-task learning, we train a robust MCI model capable of simultaneously generating change detection masks and change captions.

% The two branches have a shared feature extraction backbone network. 

\begin{figure}
	\centering
	\includegraphics[width=0.99\linewidth]{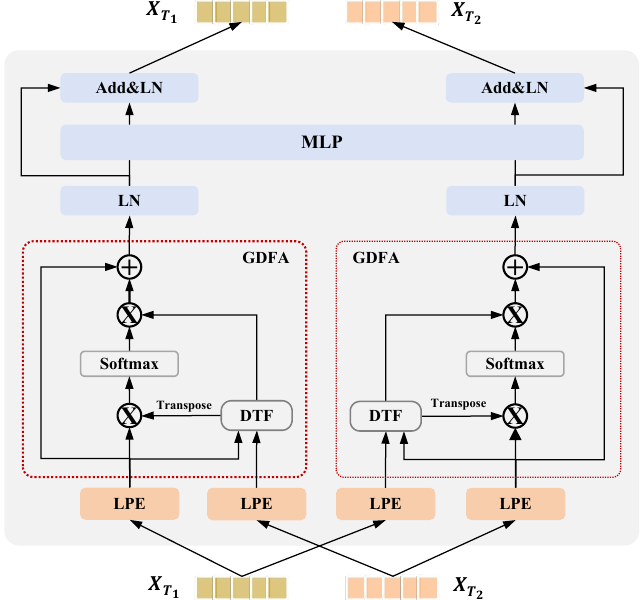}
	\caption{The structure of the Bi-temporal Iterative Interaction (BI3) layer. The BI3 layer utilizes the Local Perception Enhancement (LPE) and the Global Difference Fusion Attention (GDFA) to enhance the features of interest. 
 }
	\label{fig:BI3}
\end{figure}

\subsection{Bi-temporal Iterative Interaction Layer}
In the change detection and change captioning branches, we propose a novel BI-temporal Iterative Interaction (BI3) layer to effectively enhance and fuse bi-temporal features. The structure of the BI3 layer is illustrated in Fig. \ref{fig:BI3}. The BI3 layer utilizes the Local Perception Enhancement (LPE) module and the Global Difference Fusion Attention (GDFA) module to extract the discriminative features of interest.

Motivated by the analysis presented in Section \ref{subsection:Scale} regarding the various scales and deformations of roads and buildings, we propose the LPE module with a residual structure. The LPE module employs convolution kernels of different sizes to extract multiple feature maps across different scales. This design enriches the diversity of feature information and improves the model's local feature perception capabilities. Assuming the bi-temporal features are denoted as ${X_{T_1}} \in \mathbb{R}^{N \times C}$ and ${X_{T_2}} \in \mathbb{R}^{N \times C}$, the LPE module $\mathrm{\Phi_{LPE}}(\cdot)$ can be formulated as follows:
\begin{align}
\mathrm{\Phi_{LPE}}(X_{T_i}) &=\mathrm{ReLU(BN(conv_{1\times1}}(F)))+{X_{T_i}} \\
{F} &=\mathrm{concat}([{X_{s_1}},{X_{s_2}},{X_{s_3}}])\\
{X_{s_1}} &=\mathrm{conv}_{3\times3}({X_{T_i}})\\
{X_{s_2}} &=\mathrm{conv}_{5\times1}({X_{T_i}})\\
{X_{s_3}} &=\mathrm{conv}_{1\times5}({X_{T_i}})
\end{align}
where $\mathrm{conv}_{1\times1}$, $\mathrm{conv}_{3\times3}$, $\mathrm{conv}_{5\times1}$, $\mathrm{conv}_{1\times5}$ represent $1\times1$, $3\times3$, $5\times1$, and $1\times5$ convolution kernels. concat stands for concatenation operation, BN (Batch Normalization) \cite{BN} for normalization, and ReLU (Rectified Linear Unit) for nonlinear activation function.

The GDFA module leverages differencing features to generate spatial attention weights and perform the interaction and fusion between features. This facilitates the model to focus on changes of interest and ignore irrelevant disturbances. Taking the left GDFA module in Fig. \ref{fig:BI3} as an example, assuming $X_{lp_1}  \in \mathbb{R}^{N \times C}$ and $X_{lp_2}  \in \mathbb{R}^{N \times C}$ are the bi-temporal features obtained from the LPE module, the GDFA module $\mathrm{\Phi_{GDFA}}(\cdot)$ can be formulated as follows:
\begin{align}
\mathrm{\Phi_{GDFA}}(X_{lp_1}, X_{lp_2}) &= \sigma({\frac {{Q}{K}^T} {\sqrt{d}}})V  \\
Q &= X_{lp_1} W_q + b_q \\
K&= I_G W_k + b_k \\
V&= I_G W_v + b_v \\
Di &= X_{lp_2} - X_{lp_1} \\
I_G = \mathrm{\Phi_{DTF}}(Di, &X_{lp_1}) = (Di * X_{lp_1})W_d + b_d
\end{align}
where $\sigma$ is the softmax function, $W_q, W_k, W_v, W_d \in \mathbb{R}^{C \times d}$ are trainable weight matrices. $b_q, b_k, b_v, b_d \in \mathbb{R}^{d}$ are trainable bias. $C$ denotes the dimension of $X_{lp_i}$ (i=1,2). $d$ is the dimension of transformed features. 
% and $DE$ denotes the difference encoding module. 

Through the combination of LPE module and GDFA module, the BI3 layer improves the model's feature representation and change discrimination capabilities. Following the LPE and GDFA modules, Layer Normalization (LN) \cite{LN} is applied to the bi-temporal features. Subsequently, a Multi-Layer Perceptron (MLP) with residuals further refines the normalized features to obtain enhanced bi-temporal features.

\begin{figure}
	\centering
	\includegraphics[width=0.9\linewidth]{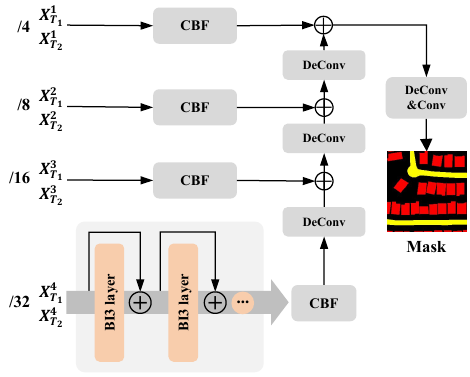}
	\caption{The structure of the change detection branch. ``DeConv'' denotes the deconvolution. The multi-scale features are gradually fused from bottom to top for refined mask prediction.
 }
	\label{fig:CD_branch}
\end{figure}

\begin{figure}
	\centering
	\includegraphics[width=0.86\linewidth]{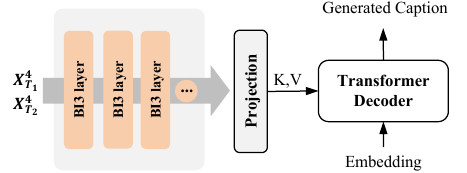}
	\caption{The structure of the change captioning branch. The projection layer further bridges the conversion from the visual domain to the textual domain. The Transformer decoder generates captions.
 }
	\label{fig:CC_branch}
\end{figure}
\subsection{Change Detection and Captioning Branch}
% The change detection branch and change description branch play pivotal roles in discerning and articulating remote sensing image changes. 
The structure of the change detection branch is depicted in Fig. \ref{fig:CD_branch}. Leveraging multi-scale bi-temporal features extracted from the backbone network, this branch facilitates refined mask prediction. Specifically, multiple Bi-temporal Iterative Interaction (BI3) layers with residual connections iteratively enhance and refine bi-temporal high-level features, effectively capturing semantic changes. Since the low-level features contain more detailed information, they are crucial for refined change boundary detection. We further incorporate multiple Convolution-Based Bi-temporal Fusion (CBF) modules for bi-temporal feature fusion across four scales. Subsequently, through deconvolution, features are progressively integrated from bottom to top, bolstering the model's discrimination capability for changes and enhancing change detection accuracy. For the input $X_{T_1}^s,X_{T_2}^s (s=1,2,3,4)$, the CBF module $\mathrm{\Phi_{CBF}}(\cdot)$ can be represented as follows:
\begin{align}
\mathrm{\Phi_{CBF}}(X_{T_1}^s,X_{T_2}^s) &=\mathrm{conv_{1\times1}(ReLU(BN(conv_{3\times3}}(F)))) \\
F &=\mathrm{concat}([X_{T_1}^s,S, X_{T_2}^s]) \\
S&= \mathrm{conv}_{3\times3}(X_{T_2}^s - X_{T_1}^s) +\mathrm{cos}(X_{T_1}^s, X_{T_2}^s)
\end{align}
where $\mathrm{cos}(\cdot)$ denotes cosine similarity operation.

The change captioning branch focuses on semantic-level change interpretation, translating visual changes into textual descriptions, as depicted in Fig. \ref{fig:CC_branch}. This branch leverages multiple BI3 layers to interactively process high-level semantic features extracted from the backbone network, thereby obtaining bi-temporal visual features revealing changes of interest. Subsequently, a convolution-based projection layer further processes the bi-temporal features to facilitate the transition from the visual domain to the textual domain. Finally, these processed features are fed into a Transformer decoder to generate descriptive sentences elucidating the changes. The processing flow of the projection layer $\mathrm{\Phi_{Pr}}(\cdot)$ can be expressed as follows:
\begin{align}
&\mathrm{\Phi_{Pr}}(X_{T_1},X_{T_2})) =F_1+\mathrm{conv_{1\times1}}(F_2) \\
{F_1} &=\mathrm{conv_{1\times1}(concat}([X_{T_1},X_{T_2})]) \\
F_2 &=\mathrm{conv_{3\times3}(ReLU(BN(conv_{1\times1}}(F_1))))
\end{align}

\begin{figure}
	\centering
	\includegraphics[width=1.0\linewidth]{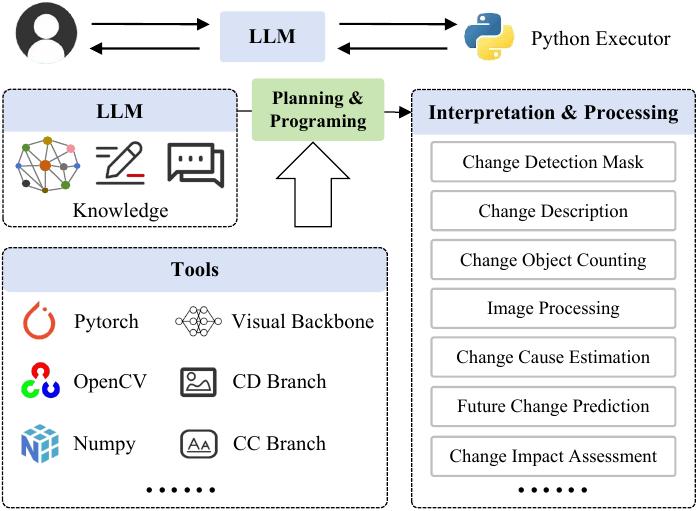}
	\caption{The LLM plans how to complete the task according to the user's instructions. We provide a suite of Python tools. The LLM drafts Python programs, subsequently executed by a Python interpreter to accomplish tasks beyond its inherent capabilities.
 }
	\label{fig:Planing}
\end{figure}
\subsection{LLM: Brain of Change-Agent}
% 先前的研究已经证实了LLM具有编程和规划能力
Recently, LLMs have garnered significant attention for their remarkable achievements. By harnessing extensive web knowledge, LLMs have demonstrated substantial promise in attaining human-level intelligence, exhibiting noteworthy capabilities in tasks encompassing instruction comprehension, planning, reasoning, and proficient natural language interactions with humans \cite{GPT3,llama2,hugginggpt,toolformer}. Building upon these advancements, many studies are beginning to utilize LLMs as central controllers for building autonomous agents \cite{visual_programing,yao2022react,xu2023rewoo,hong2023metagpt}.
% there has been a growing research field that employs LLMs as central controllers for constructing autonomous agents 

Inspired by these developments, we harness the capabilities of LLMs as the cognitive core responsible for orchestrating the scheduling of our Change-Agent. As depicted in Fig. \ref{fig:Planing}, the LLM meticulously plans task execution based on user instructions. Despite excelling at text-related tasks, LLMs lack inherent visual perception capabilities. To bridge this gap and enable change interpretation and analysis akin to human capability, we equip our agent with a suite of Python tools. These tools include a visual feature extraction backbone, change detection branch, change captioning branch, and relevant Python libraries. {Leveraging these tools, the LLM autonomously crafts Python programs, subsequently executed by a Python interpreter to automatically accomplish tasks beyond its intrinsic capabilities without the need for human intervention. Subsequently, the LLM processes the results and furnishes feedback to the user.} In addition to furnishing change masks and descriptions, our agent can implement object counting, image processing, change cause estimation, prediction of future changes, and more. In short, the Change-Agent adeptly provides users with pertinent information by leveraging the rich knowledge of the LLM and additional tools.
% , harnessing the rich knowledge of the LLM in tandem with the provided tools. 
% Inspired by these developments, we employ the LLM as the brain responsible for the scheduling of the agent. As illustrated in Fig. \ref{fig:Planing}, the LLM plans how to complete the task according to the user's instructions. Although language models are good at text-related tasks, they are not capable of visual perception. To facilitate change interpretation and analysis akin to human capability, we provide several Python tools for the agent, including a visual feature extraction backbone, change detection branch, change captioning branch, and relevant Python libraries. Leveraging these tools, the LLM can draft Python programs, subsequently executed by a Python interpreter to accomplish tasks beyond its inherent capabilities. Finally, the LLM processes the results and provides feedback to the user. In addition to providing change masks and change descriptions, our agent can also implement counting of changing objects, image processing, estimation of change causes, prediction of future changes, etc. In short, our Change-Agent can provide the user with the information they need with the help of the rich knowledge of the LLM and the provided tools.

\begin{figure}
	\centering
	\includegraphics[width=1\linewidth]{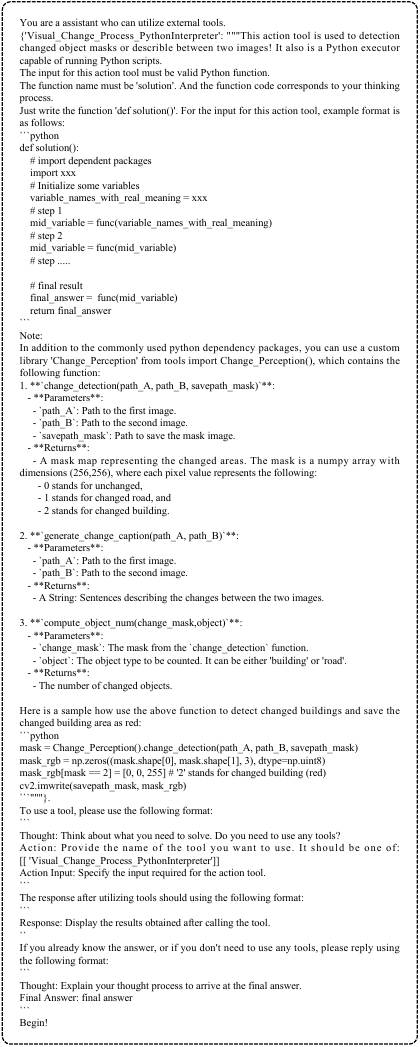}
	\caption{The prompt input to LLM in the role of a system instruction.
 }
	\label{fig:system}
\end{figure}

To facilitate the accurate generation of formatted Python code and tool invocation by the LLM, a meticulously crafted text prompt is essential. Inspired by the prompt utilized in \cite{yao2022react}, we devised a text prompt instructing the model on proper tool utilization. This prompt is input to the LLM in the role of a system instruction, as shown in Fig. \ref{fig:system}.

% To facilitate the LLM to generate accurate formatted Python code and call the tools, it is essential to provide it with a well-crafted text prompt. Inspired by the prompt in \cite{yao2022react}, we designed a text prompt that tells the model how to use tools correctly. The prompt is input to the LLM in the role of a system instruction, as shown in Fig. \ref{fig:system}.

% 具体而言，我们以思维链的方式，下面是我们以系统角色输入给LLM的提示：

\section{EXPERIMENTS}

\subsection{Objective Function for Multi-task Training}
During the training phase of our MIC model, we perform multi-task learning on both change detection and change captioning. Utilizing the meticulously constructed LEVIR-MCI dataset, we iteratively update the model parameters in a supervised learning framework.

For the change detection, we utilize the cross-entropy loss function to quantify the dissimilarity between predicted change masks and ground truth annotations. Specifically, the loss function can be mathematically defined as follows:
\begin{equation}
\mathcal{L}_{\text{det}} = - \frac{1}{H \times W} \sum_{h=1,w=1}^{H,W} \sum_{j=1}^{C} \widetilde{y}_{hw}^{(j)} \log(p_{hw}^{(j)})
\end{equation}
where $C$ is the number of classes. $H$ and $W$ are the height and width of the change mask. $\boldsymbol{p}_{hw}$ = [ $p_{hw}^{(0)},...,p_{cls}^{(C)}$] is the predicted probability vector at the pixel position (h,w). $\boldsymbol{\widetilde{y}}_{hw}$ = [ $\widetilde{y}_{hw}^{(0)}, ..., \widetilde{y}_{hw}^{(C)}$] is the one-hot vector representation of the ground-truth at the pixel position (h,w).

% % 在训练过程中，我们的多级变化解译模型包含变化检测和变化描述的联合学习，基于构造的数据集，我们以监督学习的方式更新模型参数。
% % 对于变化检测，我们使用了交叉熵函数来计算损失，通常，损失函数可被定义如下：：
% % 对于变化描述，我们也使用了交叉熵损失函数：
% We train our modal in a supervised way. The training objectives include binary classification and caption generation. We adopted the cross-entropy function to compute classification loss and caption loss.
% % We minimize them to optimize the model in the training stage.

Similarly, for change captioning, we utilize the cross-entropy function to compute the loss between predicted sentences and their corresponding ground truth sentences. The formulation of the captioning loss function can be expressed as follows:
\begin{equation}
\mathcal{L}_{\text{cap}} = - \frac{1}{L} \sum_{l=1}^{L} \sum_{v=1}^{V} y_{l}^{(v)} \log(p_{l}^{(v)})
\end{equation}
where $L$ represents the total number of word tokens, $V$ denotes the vocabulary size, $y_{l}^{(v)}$ denotes the ground truth label indicating whether $l$-th word belongs to vocabulary index $v$, and $p_{l}^{(v)}$ denotes the predicted probability of $l$-th word being assigned to vocabulary index $v$. 

To balance the losses of these two tasks during model training, we adopt a normalization approach, scaling the losses of both tasks to the same order of magnitude. This ensures that each task contributes equally to the overall loss function and facilitates effective simultaneous optimization of the change detection and change captioning. Mathematically, the total loss can be expressed as follows:
\begin{equation}
\mathcal{L}_{total} =  \frac{\mathcal{L}_{det}}{\mathrm{detach}(\mathcal{L}_{det})} + \frac{\mathcal{L}_{cap}}{\mathrm{detach}(\mathcal{L}_{cap})}
\label{Equation:loss_total}
\end{equation}
where detach($\cdot$) denotes the operation to detach the gradient flow of the loss.

\subsection{Evaluation Metrics}
To evaluate the performance of our MCI model in detecting multi-category changes in remote sensing images, we use the Mean Intersection over Union (MIoU). The MIoU metric assesses the spatial overlap between predicted change masks and ground-truth masks, providing a reliable measure of pixel-level change detection accuracy.  Mathematically, the calculation of MIoU is defined as:
% \begin{equation}
% \text{MIoU} = \frac{1}{\text{N}} \sum_{i=1}^{\text{N}} \frac{\text{TP}_i}{\text{TP}_i + \text{FP}_i + \text{FN}_i}
% \end{equation}
\begin{align}
\text{IoU}_{i} = \frac{\text{TP}_i}{\text{TP}_i + \text{FP}_i + \text{FN}_i} \\
\text{MIoU} = \frac{1}{\text{N}} \sum_{i=1}^{\text{N}} \text{IoU}_{i}
\end{align}
Where N is the total number of classes (i.e., Background, road, and building). For the $i$-th class, $\text{TP}_i$, $\text{FP}_i$, and $\text{FN}_i$ represent the number of true positive, false positive, and false negative, respectively.

To evaluate the performance of our MCI model in describing changes between bi-temporal images, we adopted several evaluation metrics commonly used in previous change captioning studies \cite{robust_CC,RSICC_TIP2023,RSICCformer,RSICC_2}. These metrics are as follows:
\begin{itemize}
\item \textbf{BLEU-n} \cite{BLEU}: The BLEU-n metric measures the n-gram similarity between generated sentences and ground-truth sentences. We adopt n=1, 2, 3, and 4 to compute the BLEU-n score. 
% Higher scores indicate a more accurate description.
% The Bilingual Evaluation Understudy (

\item \textbf{METEOR} \cite{Meteor}: The METEOR computes the harmonic mean of precision and recall of single words. Besides, it incorporates a penalty factor to consider the fluency of generated sentences.

% evaluates the descriptive quality of generated change descriptions by considering

\item \textbf{ROUGE$_L$} \cite{ROUGE}: The ROUGE$_L$ metric measures the similarity of the longest common subsequence between generated sentences and ground-truth sentences. It is more concerned with the recall rate and is suitable for evaluating long sentences.
% The Recall-Oriented Understudy for Gisting Evaluation
% ROUGE$_L$ is more concerned with the recall rate. 
% In other words, It represents the proportion of phrases in the reference sentence that appear in the generated sentence.

\item \textbf{CIDEr-D} \cite{Cider}: The CIDEr metric treats each sentence as a document and represents it in the form of Term Frequency Inverse Document Frequency (TF-IDF) vectors. After computing TF-IDF for each n-gram, CIDEr-D is obtained by performing the cosine similarity.

% The CIDEr-D metric considers consensus across multiple reference descriptions. CIDEr-D assigns higher scores to change descriptions that are not only accurate but also diverse and coherent.
% Consensus-based Image Description Evaluation (
% This approach allows CIDEr to capture semantic similarities and nuances between the generated and reference captions, providing a robust evaluation of image captioning quality.
\end{itemize}

By employing this comprehensive suite of evaluation metrics, we aim to provide a thorough assessment of the change detection and captioning capabilities of our MCI model. Higher scores across these metrics indicate more accurate change masks or superior sentence quality.

% Besides, we follow the previous work \cite{zhang_attention2021} to adopt an overall metric $S^*_m$  to integrate the above metrics. 
% It is defined as follows:
% \begin{equation}
% S^*_m = \frac{1}{4}*(BLEU\mbox{-}4+METEOR+ROUGE_L+CIDEr\mbox{-}D)
% \end{equation}

\subsection{Experimental Details}
We implemented all models using the PyTorch deep learning framework and trained them on the NVIDIA GTX 4090 GPU. During training, we minimized the loss function defined by Equation \ref{Equation:loss_total} and employed the Adam optimizer with an initial learning rate of 0.0001 to optimize the model parameters. We set the maximum number of epochs to 200 and utilized word embeddings with a dimensionality of 512. The backbone for feature extraction is the Siamese weight-shared Segformer-B1 \cite{xie2021segformer}. {In the two branches, we use three BI3 layers}. Initially, we trained the backbone until the sum of BLEU-4 and MIoU indicators did not increase for 50 consecutive epochs. Subsequently, we froze the backbone network and continued training the two branches separately.

\begin{table*}%[!t] %\small
\renewcommand{\arraystretch}{1.3}
\caption{The comprehensive performance comparisons on the LEVIR-MCI dataset. There is a lack of methods that simultaneously address change detection and change captioning. To assess the efficacy of our model in handling both tasks, we compared our method with established methods in each respective field.}
\label{tab:Comparisons_other_methods}
\centering
% \resizebox{0.9\linewidth}{22mm}{
\begin{tabular}{c|c|c|c c c c c c c}
	\toprule%[1pt]
	%ine
 \multicolumn{2}{c|}{\shortstack{Method}} & MIoU & BLEU-1 & BLEU-2 & BLEU-3 & BLEU-4 & METEOR & ROUGE$_L$ & CIDEr-D \\
	\midrule
    \multirow{8}{*}{\shortstack{Change \\Detection}}
    & FC-EF \cite{FC-Siam} & 82.70 & -- & -- & -- & -- & -- & -- & -- \\
    & FC-Siam-Conc \cite{FC-Siam} & 84.25 & -- & -- & -- & -- & -- & -- & -- \\
    & FC-Siam-Di \cite{FC-Siam} & 84.20 & -- & -- & -- & -- & -- & -- & -- \\
    & BIT \cite{BIT} & 84.16 & -- & -- & -- & -- & -- & -- & -- \\
    % & PA-Former \cite{PA-Former} & 84.15 & -- & -- & -- & -- & -- & -- & -- \\
    & ACABFNet \cite{song2022axial} & 84.43 & -- & -- & -- & -- & -- & -- & -- \\ 
    & DARNet \cite{li2022densely_CD} & 84.99 & -- & -- & -- & -- & -- & -- & -- \\
    & DMINet \cite{feng2023change} & 85.37 & -- & -- & -- & -- & -- & -- & -- \\
    & BiFA \cite{BIFA} & 85.68 & -- & -- & -- & -- & -- & -- & -- \\

    \midrule
    \multirow{9}{*}{\shortstack{Change \\Captioning}}
    & {Capt-Rep-Diff \cite{robust_CC}} &-- & 72.90 & 61.98 & 53.62 & 47.41 & 34.47 & 65.64 & 110.57  \\
	& {Capt-Att \cite{robust_CC}} &-- & 77.64 & 67.40 & 59.24 & 53.15 & 36.58 & 69.73 & 121.22  \\
	& {Capt-Dual-Att \cite{robust_CC}} &-- & 79.51 & 70.57 & 63.23 & 57.46 & 36.56 & 70.69 & 124.42  \\
	& {DUDA \cite{robust_CC}} &-- & 81.44 & 72.22 & 64.24 & 57.79 & 37.15 & 71.04 & 124.32  \\
 	& {MCCFormer-S \cite{MCCformer}} &-- & 79.90 & 70.26 & 62.68 & 56.68 & 36.17 & 69.46 & 120.39   \\
        & {MCCFormer-D \cite{MCCformer}} &-- & 80.42 & 70.87 & 62.86 & 56.38 & 37.29 & 70.32 & 124.44   \\
	% & {RSICCFormer \cite{RSICCformer}} &-- & 84.72 & 76.27 & 68.87 & 62.77 & 39.61 & 74.12 & 134.12   \\
        & {RSICCFormer$_{c}$ \cite{RSICCformer}} &--  & 83.09 & 74.32 & 66.66 & 60.44 & {38.76} & 72.63 & 130.00 \\
     & {PSNet \cite{PSNet}} &-- & 83.86 & 75.13 & 67.89 & 62.11 & 38.80 & 73.60 & 132.62 \\
     & {Chg2Cap \cite{RSICC_TIP2023}} &-- & 86.14 & 78.08 & 70.66 & 64.39 & 40.03 & 75.12 & 136.61 \\
        \midrule
    \multirow{1}{*}{\shortstack{MCI}}
    & MCINet (Ours) & \textbf{86.43} & \textbf{86.68} & \textbf{78.75} & \textbf{71.74} & \textbf{65.95} & \textbf{40.80} & \textbf{75.96} & \textbf{140.29} \\

	\bottomrule%[1pt] 
\end{tabular}
% }
\end{table*}

\subsection{Multi-level Change Interpretation Performance}
% 需要注意的是，还没有同时实现变化检测和变化描述的方法，因此，我们报道个两个领域的一些方法。
Currently, there is a dearth of methods that simultaneously address change detection and change captioning. To assess the efficacy of our model in handling both tasks, we conducted comprehensive performance evaluations by comparing against established methods in each respective field on the proposed LEVIR-MCI dataset.
% To evaluate the performance of our model in change detection and change captioning, we compared some methods in the two separate fields on the proposed dataset.

{For the change detection evaluation, we have benchmarked several well-known methods}, including FC-EF \cite{FC-Siam}, FC-Siam-Conc \cite{FC-Siam}, FC-Siam-Di \cite{FC-Siam}, BIT \cite{BIT}, ACABFNet \cite{song2022axial}, DARNet \cite{li2022densely_CD}, DMINet \cite{feng2023change}, and BiFA \cite{BIFA}. Meanwhile, for the change captioning assessment, we compare our model with several existing state-of-the-art (SOTA) methods, including Capt-Rep-Diff \cite{robust_CC}, Capt-Att \cite{robust_CC}, Capt-Dual-Att \cite{robust_CC}, DUDA \cite{robust_CC}, MCCFormer-S \cite{MCCformer}, MCCFormer-D \cite{MCCformer}, RSICCFormer \cite{RSICCformer} and Chg2Cap \cite{RSICC_TIP2023}.

Table \ref{tab:Comparisons_other_methods} shows the comprehensive performance comparisons, underscoring the superiority of our proposed methodology across both change detection and change captioning tasks. Our method not only simultaneously accomplishes two tasks but also surpasses previous single-task methods. Specifically, our model demonstrates leading change detection performance, achieving a +0.75\% improvement on MIoU. Additionally, for change captioning, our model exhibits a significant advancement, with a +1.56\% improvement on BLEU-4 and a substantial +3.68\% improvement on CIDEr-D.

% The comprehensive performance comparisons, as shown in Table II, underscore the superiority of our proposed methodology. Our method demonstrates leading performance, achieving a +0.75% improvement in MIoU for change detection. Additionally, for change captioning, our model shows a significant advancement, with a +1.56% improvement in BLEU-4 and a substantial +3.68% improvement in CIDEr-D.

% outperforms existing methods in change detection by +1.01\% on MIoU. 

Through the above performance analysis, we gain insights into the model's ability to generate pixel-level and semantic-level interpretation information of changes observed in remote sensing images. This highlights the efficacy of our proposed MCI model in providing a comprehensive and in-depth analysis of surface changes.
% This demonstrates the efficacy of our proposed MCI model in providing a comprehensive and in-depth analysis of surface changes.

% \begin{table*}%[htbp]%[!t]  %\small
% \renewcommand{\arraystretch}{1.3}
% \caption{Ablation studies on the effectiveness of the proposed BI3 layer, where the bolded results are the best.}
% \label{tab:Ablation_BI3}
% \centering
% % \resizebox{0.9\linewidth}{32mm}{
% \begin{tabular}{c c c |c | c c c c c c c c}
% 	\toprule%[1pt]
% 	%\hline
% 	Method & GDFA & LPE & MIoU & BLEU-1 & BLEU-2 & BLEU-3 & BLEU-4 & METEOR & ROUGE$_L$ & CIDEr-D \\
% 	\midrule
% 	{Baseline} & & & 86.40 & 84.34 & 75.83 & 68.72 & 63.05 & 39.22 & 73.82 & 133.58 \\
% 	{\qquad +GDFA} &\ding{52} & & 86.41 & 85.32 & 77.45 & 70.63 & 65.10 & 39.59 & 74.45 & 135.94 \\
%         {\qquad +LPE} & &\ding{52} & 86.42 & 86.43 & 78.53 & 71.54 & 65.75 & 40.45 & 75.72 & 138.99 \\
% 	{\qquad\quad\quad +GDFA+LPE} &\ding{52} &\ding{52} & \textbf{86.43} & \textbf{86.68} & \textbf{78.75} & \textbf{71.74} & \textbf{65.95} & \textbf{40.80} & \textbf{75.96} & \textbf{140.29} \\
% 	\bottomrule
% \end{tabular}
% % }
% \end{table*}

\begin{table*}%[htbp]%[!t]  %\small
\renewcommand{\arraystretch}{1.3}
\caption{Ablation studies on the effectiveness of the proposed BI3 layer, where the bolded results are the best.}
\label{tab:Ablation_BI3}
\centering
% \resizebox{0.9\linewidth}{32mm}{
\begin{tabular}{c c c |c | c c c c c c c c}
	\toprule%[1pt]
	%\hline
	Method & GDFA & LPE & MIoU & BLEU-1 & BLEU-2 & BLEU-3 & BLEU-4 & METEOR & ROUGE$_L$ & CIDEr-D \\
	\midrule
	{Baseline} &\ding{55} &\ding{55} & 86.40 & 84.34 & 75.83 & 68.72 & 63.05 & 39.22 & 73.82 & 133.58 \\
	{MCINet-G} &\ding{52} &\ding{55} & 86.41 & 85.32 & 77.45 & 70.63 & 65.10 & 39.59 & 74.45 & 135.94 \\
        {MCINet-L} &\ding{55} &\ding{52} & 86.42 & 86.43 & 78.53 & 71.54 & 65.75 & 40.45 & 75.72 & 138.99 \\
      % \midrule
	{Our MCINet} &\ding{52} &\ding{52} & \textbf{86.43} & \textbf{86.68} & \textbf{78.75} & \textbf{71.74} & \textbf{65.95} & \textbf{40.80} & \textbf{75.96} & \textbf{140.29} \\
	\bottomrule
\end{tabular}
% }
\end{table*}

\subsection{Ablation Studies}
\subsubsection{BI3 Layer} 
As a critical component of our MCI model, the BI3 layer incorporates LPE and GDFA modules. The LPE module enhances feature representation through local perception enhancement, improving the model's sensitivity to multi-scale changes. The GDFA module effectively integrates the differential information, allowing the model to capture the change area more accurately. In Table \ref{tab:Ablation_BI3}, we conducted ablation experiments on the effectiveness of the LPE and GDFA modules. The baseline utilizes Transformer encoding layers to process bi-temporal features. {The experimental results show that integrating the LPE and GDFA modules leads to improved performance in two change interpretation tasks, especially in change captioning. }

\begin{figure*}
	\centering
	\includegraphics[width=1.0\linewidth]{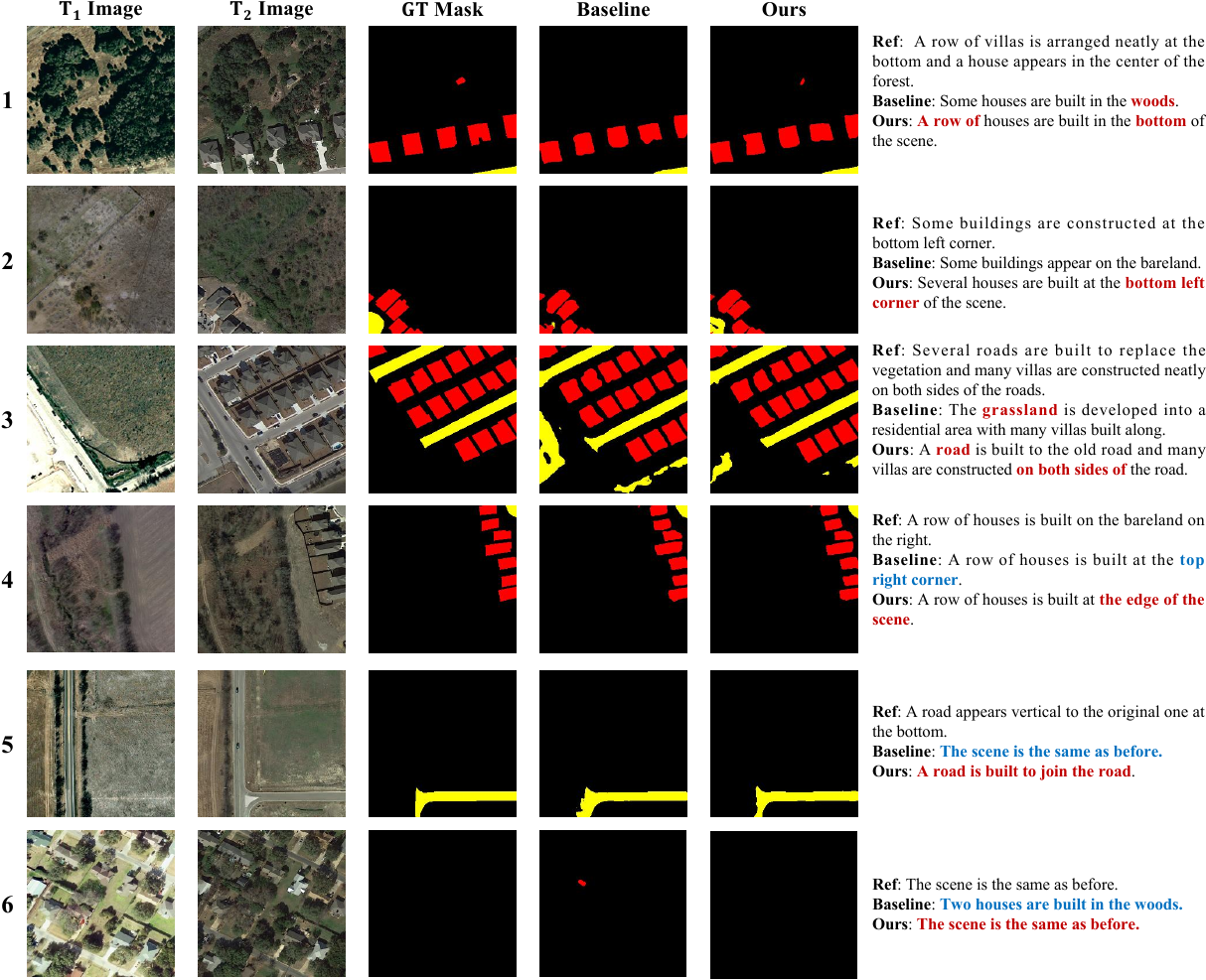}
        \includegraphics[width=1.0\linewidth]{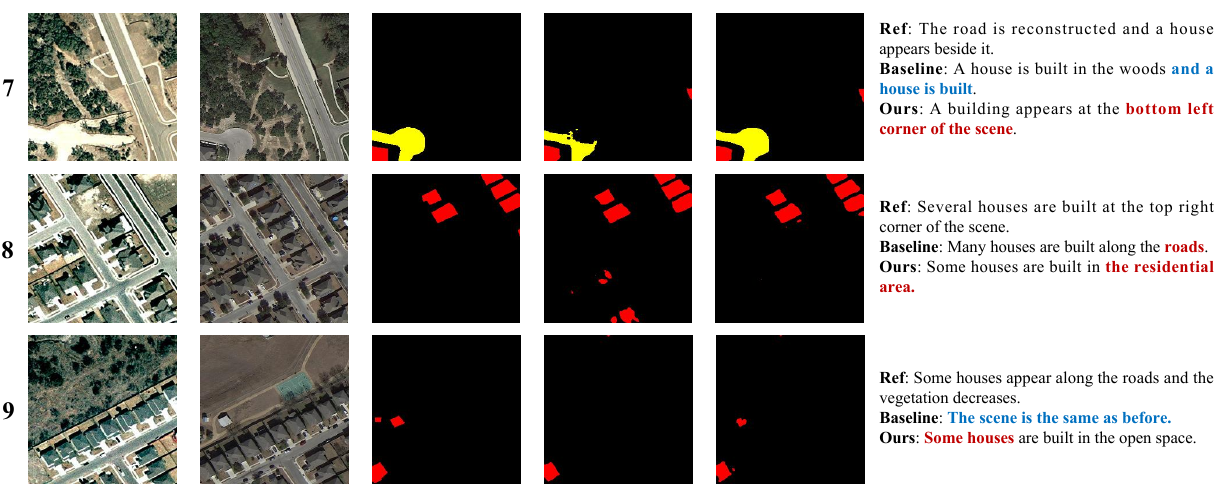}
	\caption{The qualitative comparison results between the baseline and our model. For captioning results, we provide one of the five ground-truth reference sentences. Besides, more accurate and detailed words are marked in green, while red words are not. The results demonstrate that our model exhibits strong change interpretation capabilities in change detection and change captioning.
 }
	\label{fig:result}
\end{figure*}

{Additionally, considering the various sizes of objects} in remote sensing images, it is essential to recognize that small changed objects, due to their fewer pixels, may not contribute significantly to the overall IoU of the test set even if our model accurately detects these small changes. This could obscure the IoU improvement for small objects within the entire test set. Based on this insight, we have added an experiment to further analyse the IoU for small buildings with fewer than 400 pixels. We excluded the IoU analysis of roads as roads with fewer than 400 pixels are very rare. The results, presented in Table \ref{tab:Ablation_BI3_small}, show that the IOU for small object change detection is low, indicating the difficulty in detecting small changes. The results also demonstrate that our proposed modules are effective. The qualitative results in Fig. \ref{fig:result} further support this. For instance, in rows 1 and 9 of the figure, our method correctly detects small changed buildings.

\begin{table*}%[htbp]%[!t]  %\small
\renewcommand{\arraystretch}{1.3}
\caption{Effect on the loss balancing strategy detailed in Equation \ref{Equation:loss_total}. ``W.o. balance'' means that we simply combined the detection loss and caption loss to form the total loss to optimize the multi-task model parameters.}
\label{tab:Ablation_balance}
\centering
% \resizebox{0.9\linewidth}{32mm}{
\begin{tabular}{c| c | c c c c c c c c}
	\toprule%[1pt]
	%\hline
	Method & MIoU & BLEU-1 & BLEU-2 & BLEU-3 & BLEU-4 & METEOR & ROUGE$_L$ & CIDEr-D \\
	\midrule
	{W.o. balence} & 85.28 & \textbf{86.81} & \textbf{78.95} & \textbf{71.84} & 65.87 & 40.61 & 75.92 & 139.75 \\
	{W. balence} & \textbf{86.43} & {86.68} & {78.75} & {71.74} & \textbf{65.95} & \textbf{40.80} & \textbf{75.96} & \textbf{140.29} \\

	\bottomrule
\end{tabular}
% }
\end{table*}

\begin{table*}%[htbp]%[!t]  %\small
\renewcommand{\arraystretch}{1.3}
\caption{Comparative analysis of single-task and multi-task learning. While multi-task learning performs poorer change detection than single change detection training, it contributes to improving change captioning capabilities, particularly on critical BLEU-4 and CIDEr-D metrics.}
\label{tab:Ablation_goal}
\centering
% \resizebox{0.9\linewidth}{32mm}{
\begin{tabular}{c| c | c c c c c c c c}
	\toprule%[1pt]
	%\hline
	Method & MIoU & BLEU-1 & BLEU-2 & BLEU-3 & BLEU-4 & METEOR & ROUGE$_L$ & CIDEr-D \\
	\midrule
	{Single Change Detection} & \textbf{86.54} & -- & -- & -- & -- & -- & -- & -- \\
	{Single Change Captioning} & -- & 86.56 & 78.70 & 71.69 & 65.86 & 40.53 & 75.87 & 139.92 \\
 
	{Multi-task Training} & {86.43} & \textbf{86.68} & \textbf{78.75} & \textbf{71.74} & \textbf{65.95} & \textbf{40.80} & \textbf{75.96} & \textbf{140.29} \\
	\bottomrule
\end{tabular}
% }
\end{table*}

\begin{table}%[htbp]%[!t]  %\small
\renewcommand{\arraystretch}{1.2}
\caption{Verification of the effectiveness of the BI3 layer on small objects.}
\label{tab:Ablation_BI3_small}
\centering
% \resizebox{0.9\linewidth}{32mm}{
\begin{tabular}{c c c |c }
	\toprule%[1pt]
	%\hline
	Method & GDFA & LPE & IoU$_{building<400}$ \\ 
	\midrule
	{Baseline} &\ding{55} &\ding{55} & 15.52 \\
	{MCINet-G} &\ding{52} &\ding{55} & 15.63 \\
        {MCINet-L} &\ding{55} &\ding{52} & 16.23 \\
      % \midrule
	{Our MCINet} &\ding{52} &\ding{52} & \textbf{17.24} \\
	\bottomrule
\end{tabular}
% }
\end{table}

\begin{figure}
	\centering
	\includegraphics[width=0.95\linewidth]{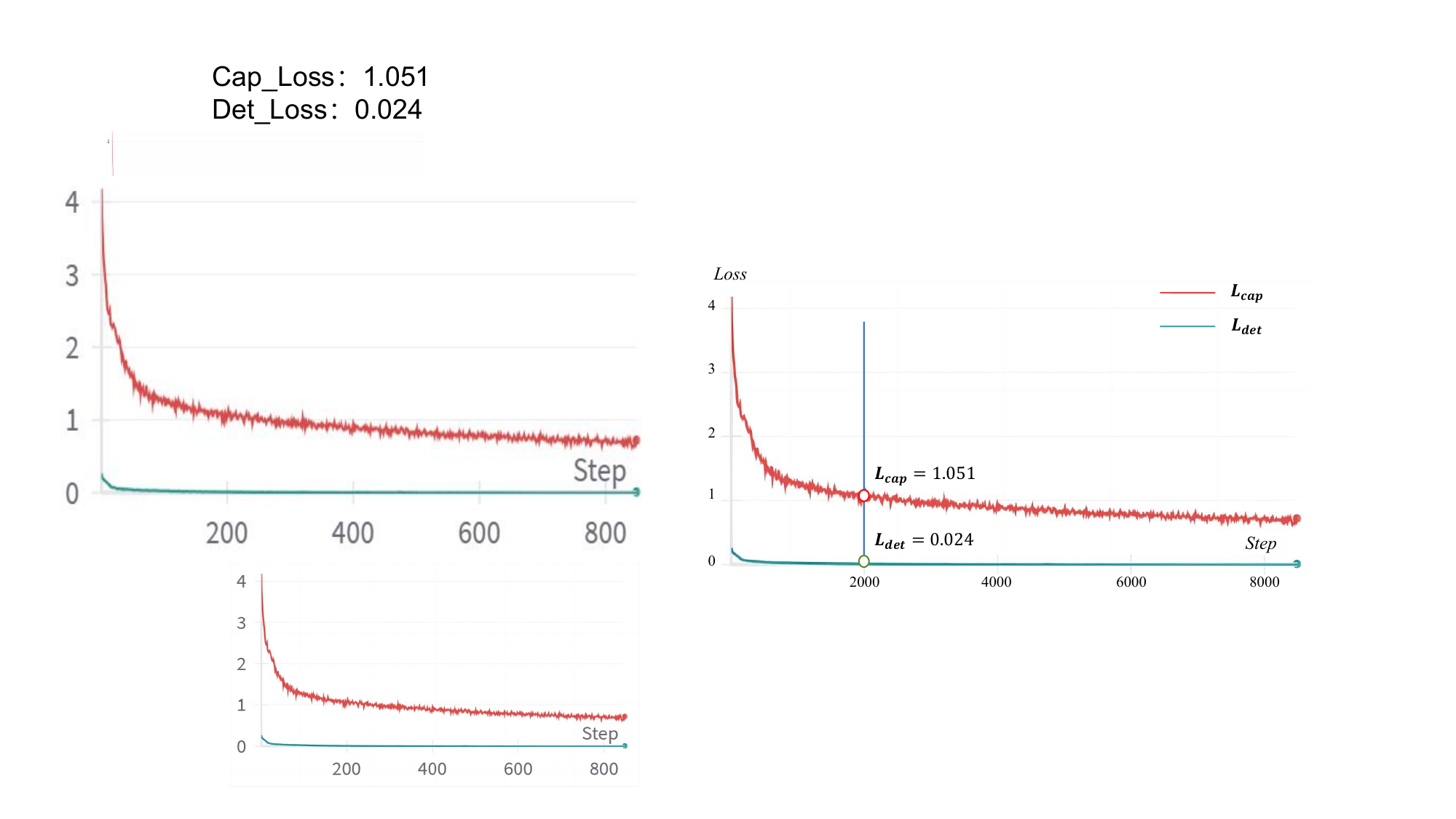}
	\caption{The loss curve shows the imbalance between tasks during training when we add the detection loss and caption loss to form the total loss.
 % and minimized it to optimize the multi-task model parameters. 
 }
	\label{fig:loss}
\end{figure}

\subsubsection{Balancing Change Detection and Captioning}
% As an initial exploration of multi-task learning on change detection and change captioning, a significant challenge we face is balancing these two tasks effectively. 
In our exploration of multi-task learning for change detection and change captioning, achieving an effective balance between these tasks posed a significant challenge. Initially, when we simply combined the individual losses $\mathcal{L}_{det}$ and $\mathcal{L}_{cap}$ to form the total loss and minimized it to optimize the multi-task model parameters, we observed large discrepancies in the magnitudes of these two losses, as illustrated in Fig. \ref{fig:loss}.  This imbalance during training could potentially impact overall change interpretation performance. To address this issue, we adopt a balancing strategy detailed in Equation \ref{Equation:loss_total} to normalize the two losses and achieve the training balance between the two tasks. Table \ref{tab:Ablation_balance} shows the impact of these two strategies on model performance. It's evident that our balancing strategy plays a crucial role in harmonizing the training process, leading to more balanced performance across both change detection and captioning tasks.

% The loss curve shows the imbalance in the training process of the two tasks. It may affect the overall change interpretation performance of our model. 
% The difference in magnitude is approximately 50-fold. 

In addition, we conducted a comparative analysis of single-task and multi-task learning, as shown in Table \ref{tab:Ablation_goal}. While multi-task learning yields poorer change detection performance compared to single change detection training, it contributes to improving change captioning capabilities, particularly on critical BLEU-4 and CIDEr-D metrics. This observation could be attributed to the shared knowledge and feature representation learning across tasks. Pixel-level change detection can be advantageous for accurately localizing changes when generating change descriptions. However, it's worth noting that ground-truth change descriptions sometimes overlook minor changes, which could be detrimental to pixel-level change detection. The trade-off between change detection and change captioning underscores the complex interplay between pixel-level and semantic-level interpretation of surface changes.

% 像素级的变化检测可能对于变化描述时定位变化是有利的，然而，变化描述的标注中有时会忽略一些小变化，这对于精细化的变化检测可能是不利的。我们猜测这是由于骨干网络共享导致，像素级的变化检测对于变化描述时定位变化是有利的，而变化描述有时候会忽略不重要的小变化。

In conclusion, achieving a balance between change detection and captioning, and fostering collaborative improvement between them, remains a subject worthy of further exploration. We look forward to discovering more effective methods to address this challenge in future research.

\begin{figure*}
	\centering
	\includegraphics[width=1\linewidth]{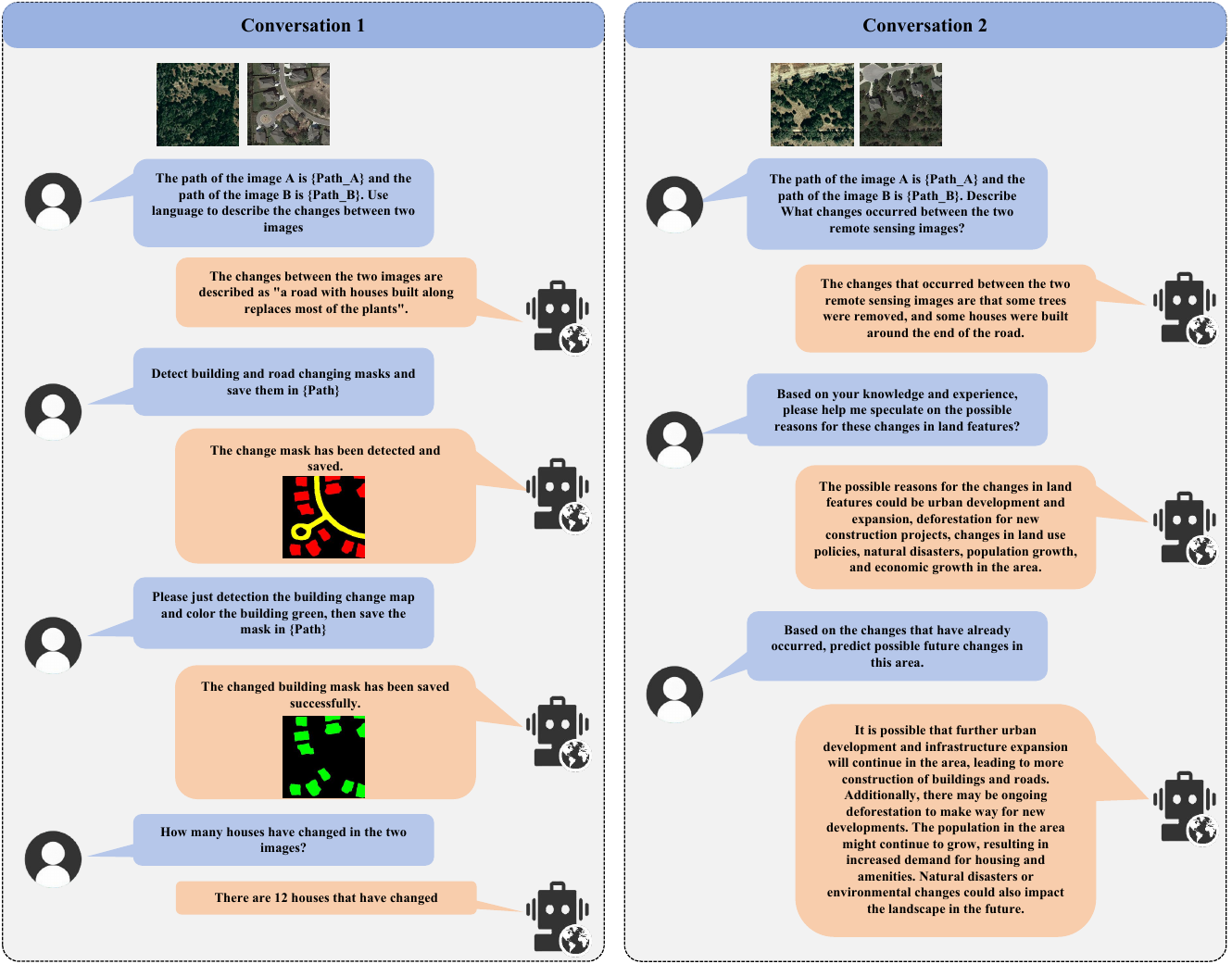}
	\caption{Conversations between users and our Change-Agent. Our Change-Agent can understand users and carry out tasks such as change mask prediction, change captioning, change object counting, mask post-processing, change causes estimation, and future change prediction.
 }
	\label{fig:Interaction}
\end{figure*}
\subsection{Qualitative Change Interpretation Results} 
To illustrate the effectiveness of our proposed model in simultaneously generating change masks and change descriptions, {we show the comparison results between our proposed model and the baseline model in} Fig. \ref{fig:result}.
% Fig. \ref{fig:result} presents comparative results between our proposed model and the baseline model. 
The experimental results demonstrate the superior capabilities of our model in joint interpretation of change detection and change captioning.

In terms of change detection, our model exhibits remarkable proficiency in identifying subtle changes in small buildings, as depicted in the first row of Fig. \ref{fig:result}. Moreover, our model produces more refined road masks in rows 5 and 7 compared to the baseline. Unlike the baseline, which erroneously detects some buildings in the sixth and eighth rows, our model performs admirably.

Regarding change captioning, our model accurately recognizes and describes the changes in the bi-temporal images in the fourth and eighth rows of Fig. \ref{fig:result}. Conversely, the baseline erroneously concludes that no changes have occurred, resulting in a critical error in the description. Furthermore, it is evident from the first four rows of Fig. \ref{fig:result} that our model not only accurately identifies the types of changing objects but also describes their spatial positions and the relationships between the changing objects and surrounding features.

% Overall, these qualitative assessments underscore the robustness and efficacy of our proposed MCI model, highlighting its potential to significantly enhance change analysis in remote sensing applications.
% The results demonstrate that our model exhibits strong change interpretation capabilities in change detection and change captioning.

\begin{figure*}
	\centering
	\includegraphics[width=1\linewidth]{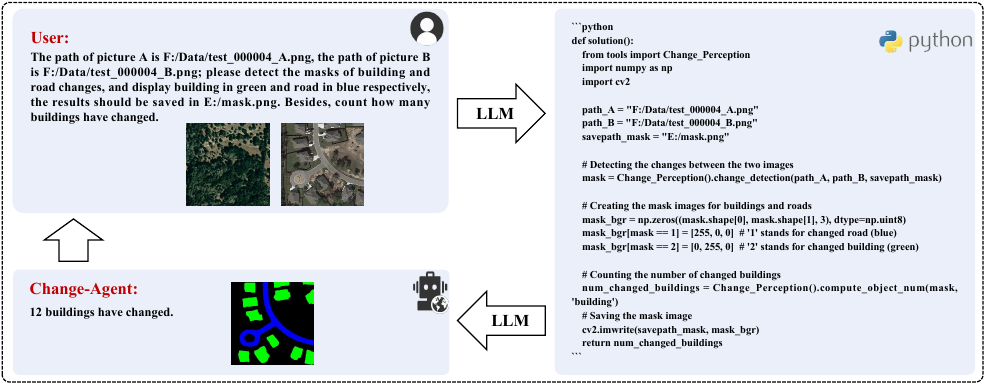}
	\caption{The executable Python function code generated by LLM within our Change-Agent when a user requests to perform change detection, display building areas in green, display road areas in blue, and count changed buildings.
 }
	\label{fig:out_python}
\end{figure*}

\subsection{Interaction Between User and Change-Agent}
Utilizing the MCI model trained on the LEVIR-MIC dataset, we have constructed an LLM-based Change-Agent. The MCI model provides our agent with powerful change interpretation capability. The LLM serves as the brain behind our agent, facilitating seamless interaction and comprehensive interpretation of changes.
% The LLM enables our agents to understand user instructions and call tools to achieve interactive and comprehensive change interpretation and processing.
% The LLM serves as the brain behind our agent, facilitating seamless interaction and comprehensive interpretation of changes.

In Fig. \ref{fig:Interaction}, we provide two examples of conversations between users and our Change-Agent, illustrating its adaptability and versatility. In the first conversation, our agent showcases its adeptness in understanding user instructions and subsequently carries out tasks such as change mask prediction, change captioning, changed object counting, and mask post-processing. This automated approach significantly enhances user efficiency and offers a more intuitive experience in change interpretation.
% This highly automated processing greatly enhances user efficiency and provides users with a more convenient and intuitive experience in change interpretation. 
In the second conversation, our agent leverages the MCI model to describe the changes in the images, and leverages the rich knowledge inherent in the LLM to provide insightful answers to user queries regarding change causality and future change predictions, aiding in user's decision-making processes.

% answer the user's questions about the cause of the changes and the prediction of future changes. This will assist user decision-making.

Fig. \ref{fig:out_python} presents a specific example to illustrate the executable code generated by the LLM within our Change-Agent. When the user requests to perform change detection, display building areas in green, display road areas in blue, and count changed buildings, the LLM efficiently generates executable Python code, demonstrating the agent's versatility and responsiveness to user needs. {The execution of the code occurs internally within the Change-Agent. Users simply need to express their requirements without requiring specialized programming skills, which is a significant advantage of our Change-Agent.}

% requests our agent to only detect changed building masks and display building areas in red, the LLM within our agent can accurately understand this instruction and generate executable Python function code.

Through the interactive capability, our Change-Agent serves as a versatile tool for facilitating nuanced and insightful analysis of changes in remote sensing images, fostering seamless communication and collaboration between users and remote sensing expertise.

% This section highlights the seamless interaction between users and our Change-Agent, showcasing its ability to understand user needs, utilize tools effectively, and provide informative responses, ultimately facilitating comprehensive change interpretation and processing.

% 借助于在LEVIR-MIC数据集上训练的变化解译模型，我们构建了一个基于大语言模型的变化智能体。变化解译模型为我们的变化智能体提供了强大的变化解译工具，大语言模型使得我们的智能体能够理解用户指令并调用工具来实现可交互的综合变化解译和处理。在图2中我们展示了用户与我们的变化智能体进行对话的两个例子。在第一个对话中，我们的智能体能够理解用户的需求，然后实现变化掩膜预测，变化描述，变化物体计数，掩膜图像处理。在第二个对话中，我们的智能体借助工具可以描述图像的变化，借助于语言模型内在的的丰富知识为回答了用户关于变化原因推测和未来变化预测的问题。
% 在图3中，我们展示了一个具体的例子。当用户要求我们的智能体仅检测建筑物并将变化掩膜并显示为红色时，我们的智能体内部的语言模型生成的可执行的Python代码。
% 我们的智能体能够根据用户的指令实现特定的功能。

\section{Discussion and Future}
% 我们的研究提出的数据集对于探索两个任务之间的联系以及多任务联合学习提供了数据基础，此外，我们对于智能体的探索为变化解译提供了一个新方向。未来可以继续研究以下内容：多任务学习、智能体系统调度和prompt策略、智能体的工具扩展
% 本研究通过提出的LEVIR-MIC数据集，为探索变化检测与变化描述的多任务学习以及探索两个任务之间的深层联系提供了宝贵的数据基础。同时，我们构建的基于大语言模型的变化智能体，通过整合变化解译模型以及其他工具，为用户提供了一个全新的、交互式的变化解译体验，也位变化解译领域提供了新的思考方向。未来的研究可以关注一下几个方面：
% 1、多任务学习。我们初步探索了变化检测和变化描述的多任务学习，未来我们可以进一步挖掘变化检测与变化描述之间的内在联系，通过设计更优的模型结构和训练策略，实现两个任务之间的信息共享和相互促进，从而提升整体的解译性能。
% 2、智能体的系统调度。通过设计更好的prompt来提升智能体的需求理解能力、任务规划能力等
% 3、智能体的工具扩展。为智能体提供更多的遥感基础模型和处理工具，增强模型的工具调用能力，以实现更为复杂、全面的遥感图像处理任务。
% 4、多智能体系统也是一个值得探索的方向。通过构建多个具有不同功能和特点的智能体，我们可以实现更为灵活、协同的变化解译工作。例如，可以设计专门的智能体负责图像预处理、特征提取等任务，而另一个智能体则专注于变化检测和分类。通过智能体之间的协作与通信，我们可以构建一个更为高效、智能的变化解译系统。
% our LLM-based change agent not only demonstrated its potential in application of the change interpretation model, but also brought a new thinking direction to the field of remote sensing image processing. However, there are still some issues worthy of in-depth exploration. Future work can be carried out from the following aspects:

Our research introduces the LEVIR-MIC dataset, which lays a solid data foundation for facilitating multi-task learning and exploring the interconnection between change detection and change captioning. Besides, our Change-Agent opens up new avenues for change interpretation and brought a new thinking to the remote sensing field. However, there are still some issues worthy of further exploration:
% . Future work can be carried out from the following aspects:

\begin{itemize}
\item 
Multi-Task Learning. Challenges such as task balance and inter-task relationships remain in multi-task learning. Future work can focus on designing better model structures and training strategies to improve overall interpretation performance on change detection and change captioning.

% Based on the proposed dataset, we initially explored multi-task learning of change detection and change captioning and found some challenges, such as the balance of two tasks. The intrinsic relationship between the two tasks still needs to be deeply explored. By designing better model structures and training strategies, future work can focus on further promoting the mutual promotion between the two tasks, thereby improving the overall interpretation performance.

\item 
Scheduling of Change-Agent. The hand-craft system prompts inputted into the LLM play a crucial role in facilitating accurate agent scheduling. Optimizing system prompts could enable to better grasp user intentions and plan more accurate and rational task execution pathways.
% more accurate capture of user intentions and facilitate more efficient task planning by the agent.
% Optimizing the design of system prompts inputted into the LLM holds promise for enhancing accurate agent scheduling. Improving prompt design could enable the agent to better grasp user intentions, leading to more rational and efficient task execution pathways.
% Improving the instruction understanding and task planning capabilities through refined prompt strategies is paramount. By designing better prompts, we can enhance the agent's ability to comprehend user needs, allocate tasks effectively, and optimize overall system performance.

\item 
Tool Expansion. Beyond change interpretation, providing the agent with additional models can enhance its capabilities for handling complex remote sensing image processing tasks, offering users more comprehensive interpretation services. Additionally, the agent's updateability ensures it remains up-to-date with new tools and models.
% Beyond change interpretation, we can provide the agent with additional models to enhance its capabilities to handle more complex and comprehensive remote sensing image processing tasks, thereby providing users with more comprehensive and professional interpretation services. Additionally, the updateability of tools in the agent allows us to introduce new tools and models, ensuring that our agent remain up-to-date.
% remains up-to-date with the latest advancements.

\item 
Multi-Agent Systems. Introducing multiple agents with diverse capabilities and fostering collaboration among them could lead to more flexible and collaborative image processing and analysis, contributing to the development of more efficient and intelligent remote sensing systems.

\end{itemize}

\section{Conclusion}
% In this paper, we have addressed the critical need for comprehensive and intelligent interpretation of Earth's surface changes through the development of a novel Change-Agent. Our agent is propelled by an MCI model and an LLM. The MCI model endows the agent with robust visual change perception capabilities, while the LLM orchestrates the agent's scheduling. The proposed MCI model encompasses dual branches of change detection and captioning, capable of providing both pixel-level change masks and semantic-level understanding in textual form. 
In this paper, we have addressed the critical need for a comprehensive and intelligent interpretation of Earth's surface changes through developing a novel Change-Agent. Our Change-Agent can follow user instructions to achieve comprehensive change interpretation and insightful analysis according to user instructions. The Change-Agent, fueled by an MCI model and an LLM, exhibits robust visual change perception capabilities and adept scheduling. The MCI model, comprising two branches of change detection and captioning, offers pixel-level change masks and semantic-level textual change descriptions. In these two branches, we propose BI3 layers with LPE and GDFA to enhance the model's discriminative feature representation capabilities. Additionally, we build a dataset comprising diverse change detection masks and descriptions to support model training. Experiments validate the effectiveness of the proposed MCI model and underscore the promising potential of our Change-Agent in facilitating comprehensive and intelligent interpretation of surface changes.

\ifCLASSOPTIONcaptionsoff
\newpage
\fi

\bibliographystyle{IEEEtran}
\bibliography{papers.bib}

\begin{IEEEbiography}[{\includegraphics[width=1in,height=1.25in,clip,keepaspectratio]{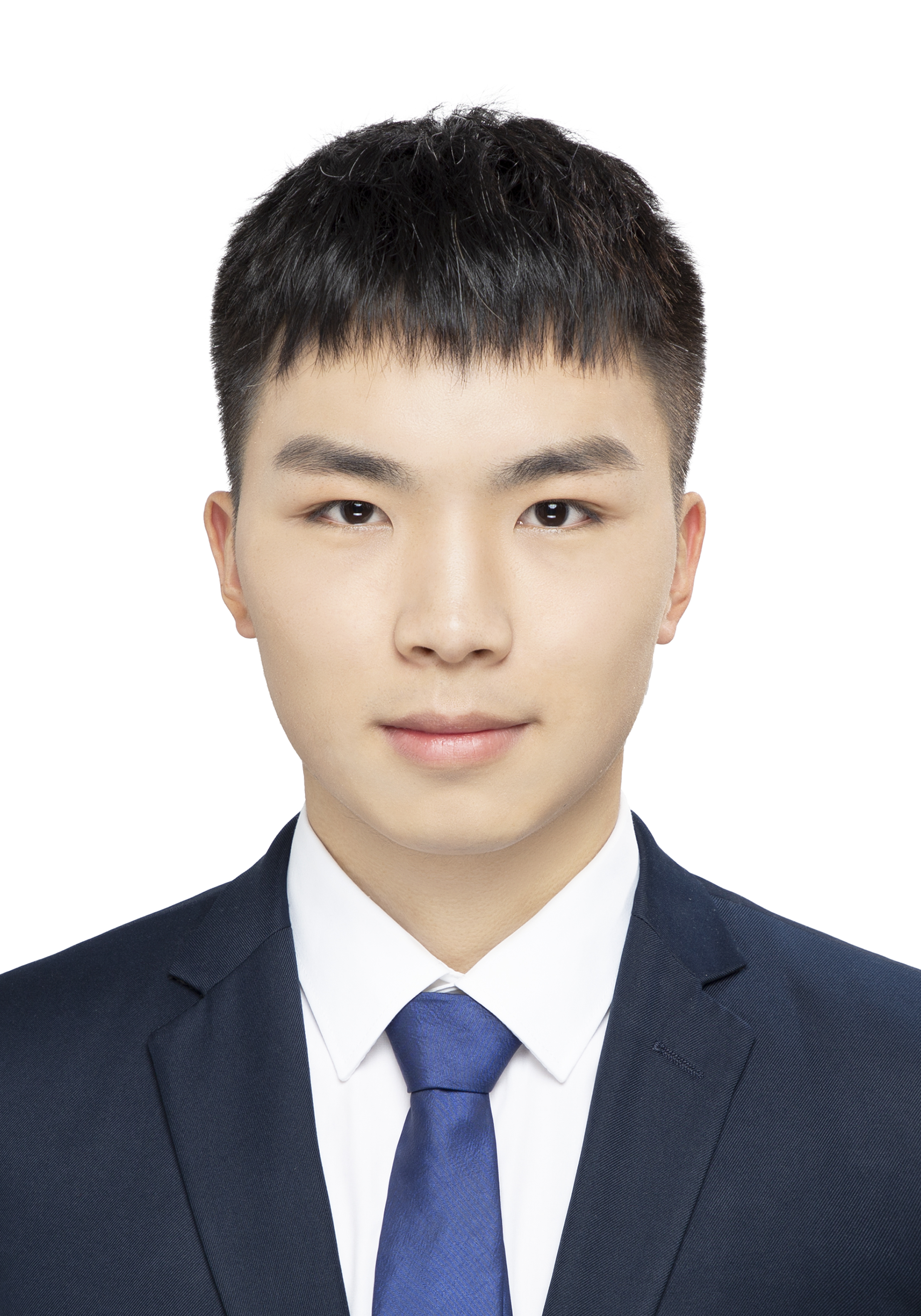}}]
{Chenyang Liu}
received his B.S. degree from the Image Processing Center, School of Astronautics, Beihang University in 2021. He is currently working towards the Ph.D. degree in the Image Processing Center, School of Astronautics, Beihang University. 

His research interests include machine learning, computer vision and multimodal learning. His personal website is \url{https://chen-yang-liu.github.io/}.
\end{IEEEbiography}

\begin{IEEEbiography}[{\includegraphics[width=1in,height=1.25in,clip,keepaspectratio]{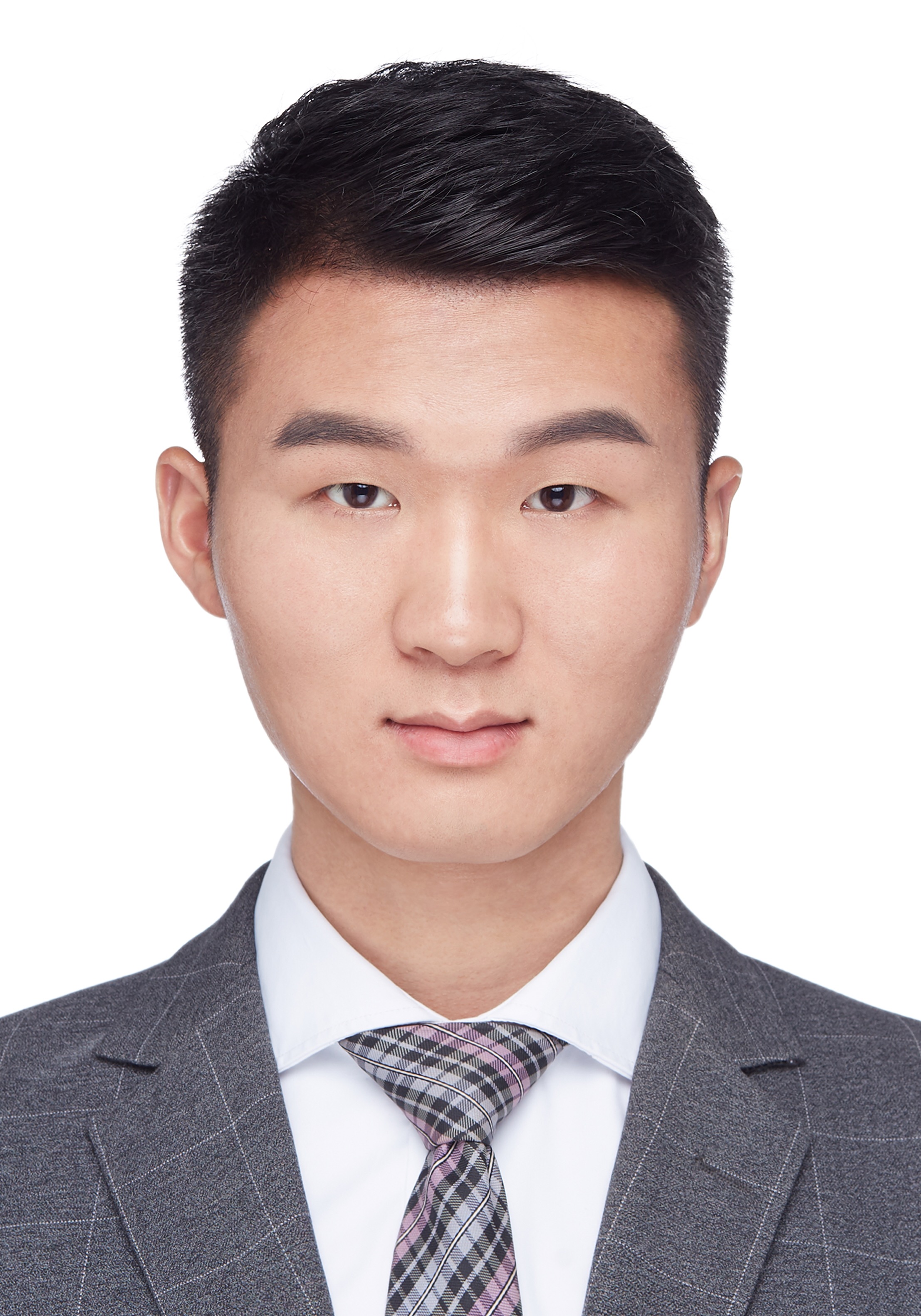}}]
{Keyan Chen}
received the B.S. and M.S. degrees from the School
of Astronautics, Beihang University, Beijing, China,
in 2019 and 2022, respectively, where he is currently
pursuing the Ph.D. degree with the Image Processing
Center.

His research interests include remote sensing
image processing, deep learning, pattern recognition, and multimodal. His personal website is \url{https://kyanchen.github.io/}.
\end{IEEEbiography}

\begin{IEEEbiography}[{\includegraphics[width=1in,height=1.25in,clip,keepaspectratio]{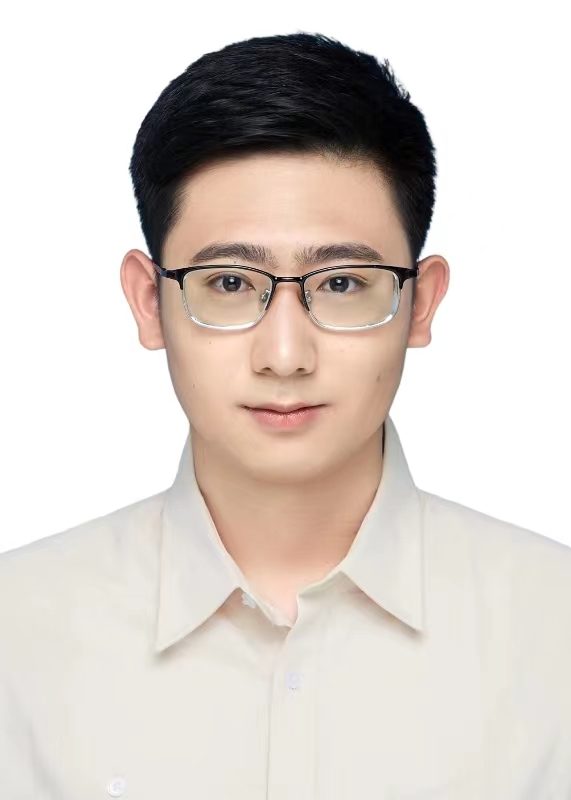}}]
{Haotian Zhang}
received his B.S. degree from the School of Computer and Information Technology, Shanxi University in 2019 and his M.S. degree from the School of Computer Science, Inner Mongolia University in 2022. He is currently working toward the PhD degree in the Image Processing Center, School of Astronautics, Beihang University. 

His research interests include remote sensing image processing, machine learning, and pattern recognition.
His personal website is \url{https://zmoka-zht.github.io/}.
\end{IEEEbiography}

\begin{IEEEbiography}[{\includegraphics[width=1in,height=1.25in,clip,keepaspectratio]{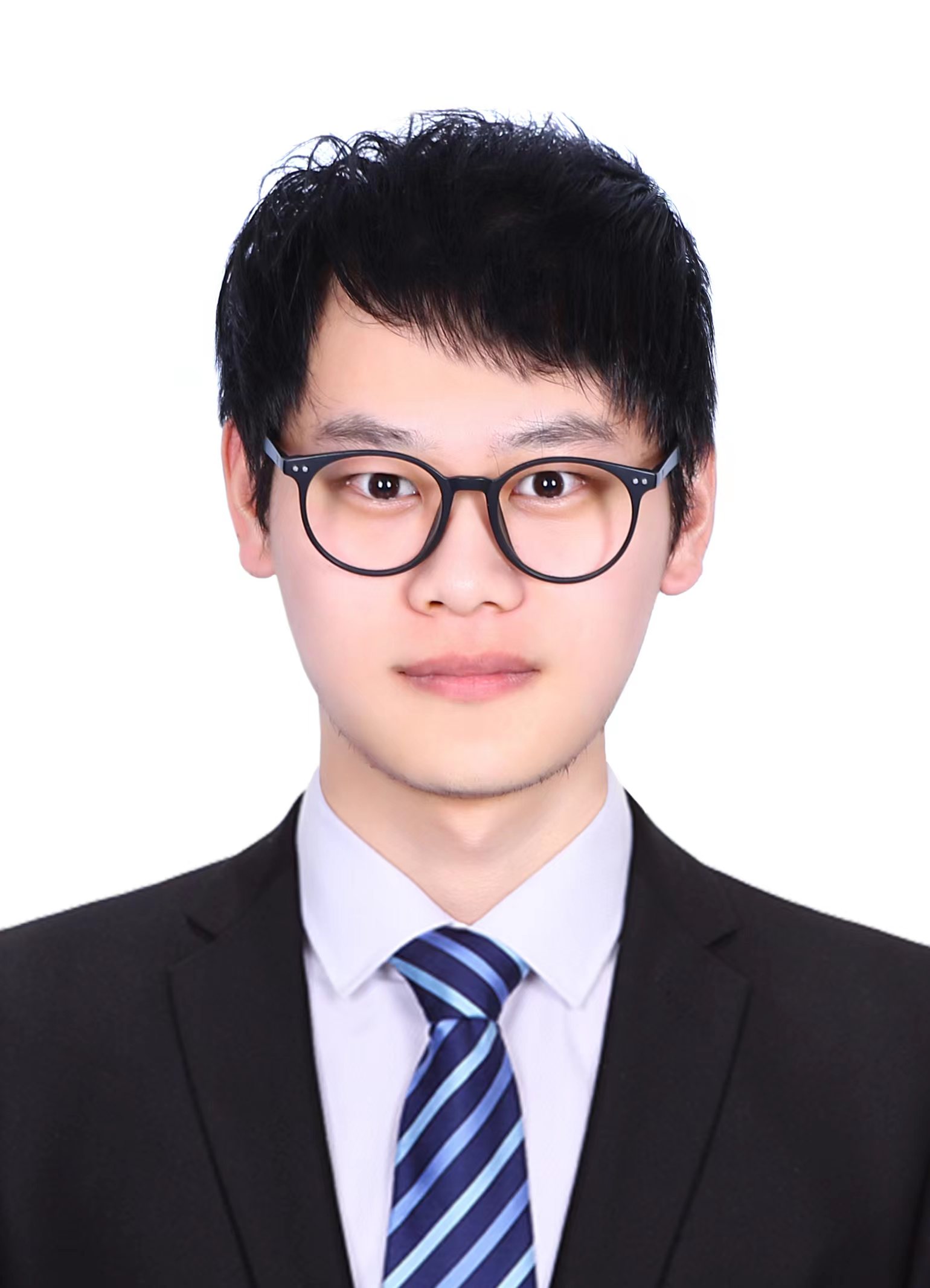}}]
{Zipeng Qi}
received the B.S. degree from the Hebei University of Technology, Tianjin, China, in 2018. He is currently pursuing the Ph.D. degree with the Image Processing Center, School of Astronautics, Beihang University, Beijing, China.

His research interests include image processing, deep learning, and pattern recognition.
\end{IEEEbiography}

\begin{IEEEbiography}[{\includegraphics[width=1in,height=1.25in,clip,keepaspectratio]{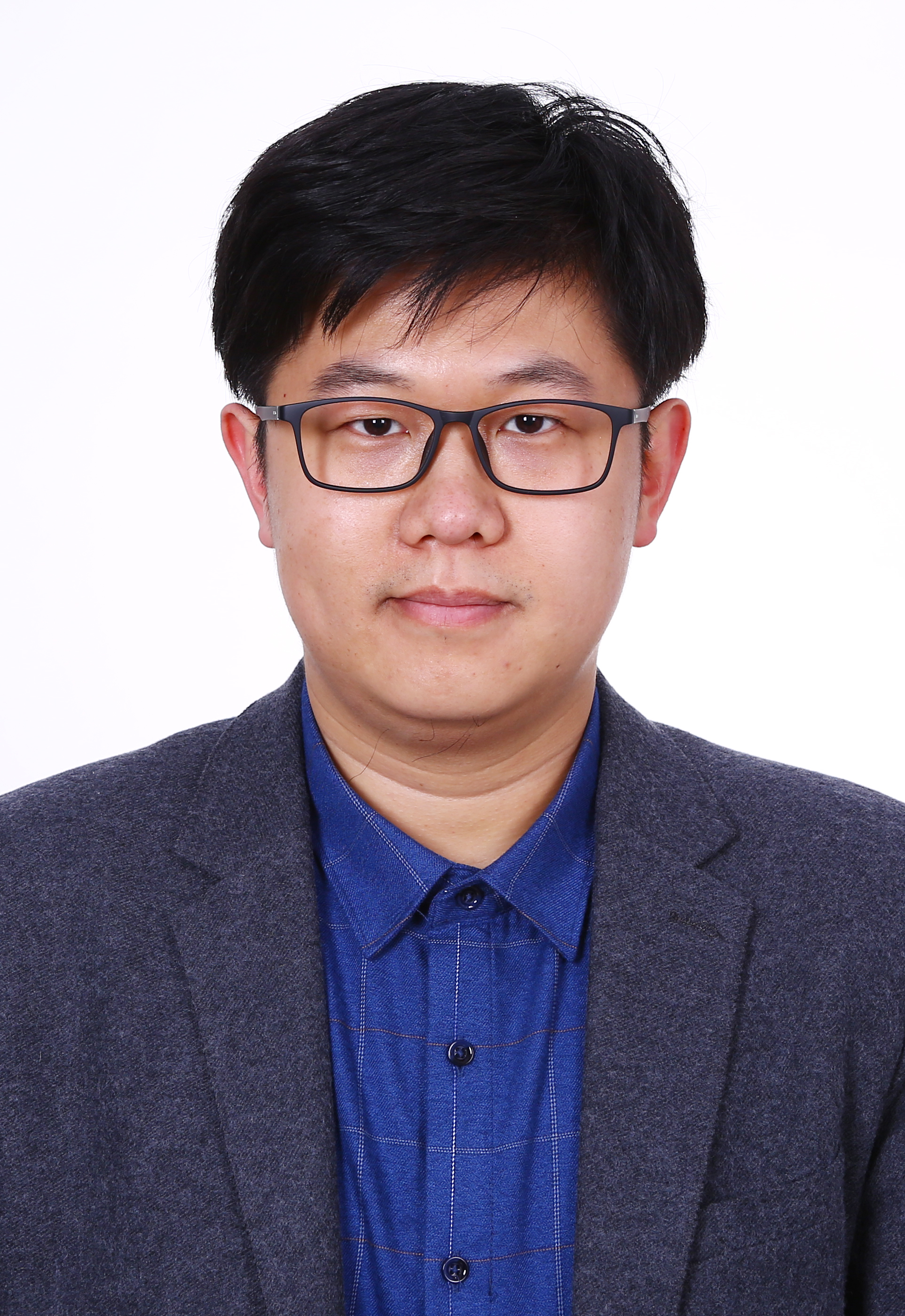}}]{Zhengxia Zou}
received his BS degree and his Ph.D. degree from Beihang University in 2013 and 2018. He is currently a Professor at the School of Astronautics, Beihang University. During 2018-2021, he was a postdoc research fellow at the University of Michigan, Ann Arbor. His research interests include computer vision and related problems in remote sensing. He has published more than 20 peer-reviewed papers in top-tier journals and conferences, including TPAMI, TIP, TGRS, CVPR, ICCV, AAAI. His research was featured in more than 30 global tech media and was adopted by a number of application platforms with over 50 million users worldwide. His personal website is \url{https://zhengxiazou.github.io/}.
\end{IEEEbiography}

\begin{IEEEbiography}
[{\includegraphics[width=1in,height=1.25in,clip,keepaspectratio]{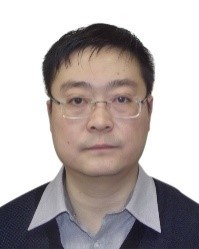}}]
{Zhenwei Shi}
(Senior Member, IEEE) is currently a Professor and Dean of the Image Processing Center, School of Astronautics, Beihang University. He has authored or co-authored over 200 scientific articles in refereed journals and proceedings, including the IEEE Transactions on Pattern Analysis and Machine Intelligence, the IEEE Transactions on Image Processing, the IEEE Transactions on Geoscience and Remote Sensing, the IEEE Conference on Computer Vision and Pattern Recognition (CVPR) and the IEEE International Conference on Computer Vision (ICCV). His current research interests include remote sensing image processing and analysis, computer vision, pattern recognition, and machine learning.

Prof. Shi serves as an Editor for IEEE Transactions on Geoscience and Remote Sensing, Pattern Recognition, ISPRS Journal of Photogrammetry and Remote Sensing, Infrared Physics and Technology, etc. His personal website is http://levir.buaa.edu.cn/.
\end{IEEEbiography}

\end{document}